\documentclass[nohyperref]{article}

\usepackage{hyperref}       
\usepackage{booktabs}       
\usepackage{amsfonts}       
\usepackage{nicefrac}       
\usepackage{pifont}        
\usepackage{microtype,amsthm}
\usepackage{subfigure}
\usepackage{amsmath}
\usepackage{amssymb}
\usepackage{bbm}
\usepackage{mathtools}
\usepackage{array,adjustbox}
\usepackage{multirow}
\usepackage{changes}

\usepackage[accepted]{icml2022}

\usepackage[capitalize,noabbrev]{cleveref}

\usepackage{natbib}
\theoremstyle{plain}
\newtheorem{theorem}{Theorem}[section]
\newtheorem{proposition}[theorem]{Proposition}
\newtheorem{lemma}[theorem]{Lemma}
\newtheorem{corollary}[theorem]{Corollary}
\newtheorem{definition}[theorem]{Definition}
\newtheorem{assumption}[theorem]{Assumption}
\newtheorem{remark}[theorem]{Remark}

\icmltitlerunning{AI-IRW depth}

\begin{document}

\twocolumn[
\icmltitle{Affine-Invariant Integrated Rank-Weighted Depth:\\ Definition, Properties and Finite Sample Analysis}



\icmlsetsymbol{equal}{*}

\begin{icmlauthorlist}
\icmlauthor{Guillaume Staerman}{yyy}
\icmlauthor{Pavlo Mozharovskyi}{yyy}
\icmlauthor{Stéphan Clémençon}{yyy}
\end{icmlauthorlist}

\icmlaffiliation{yyy}{LTCI, Télécom Paris, Institut Polytechnique de Paris}

\icmlcorrespondingauthor{Guillaume Staerman}{guillaume.staerman@telecom-paris.fr}

\icmlkeywords{Machine Learning, ICML}

\vskip 0.3in
]



\printAffiliationsAndNotice{\icmlEqualContribution} 

\begin{abstract}
Because it determines a center-outward ordering of observations in $\mathbb{R}^d$ with $d\geq 2$, the concept of statistical depth permits to define quantiles and ranks for multivariate data and use them for various statistical tasks (e.g. inference, hypothesis testing). Whereas many depth functions have been proposed \textit{ad-hoc} in the literature since the seminal contribution of \citet{Tukey75}, not all of them possess the properties desirable to emulate the notion of quantile function for univariate probability distributions. In this paper, we propose an extension of the \textit{integrated rank-weighted} statistical depth (IRW depth in abbreviated form) originally introduced in \citet{IRW}, modified in order to satisfy the property of \textit{affine-invariance}, fulfilling thus all the four key axioms listed in the nomenclature elaborated by \citet{ZuoS00a}. The variant we propose, referred to as the Affine-Invariant IRW depth (AI-IRW in short), involves the precision matrix of the (supposedly square integrable) $d$-dimensional random vector $X$ under study, in order to take into account the directions along which $X$ is most variable to assign a depth value to any point $x\in \mathbb{R}^d$. The accuracy of the sampling version of the AI-IRW depth is investigated from a non-asymptotic perspective. Namely, a concentration result for the statistical counterpart of the AI-IRW depth is proved. Beyond the theoretical analysis carried out, applications to anomaly detection are considered and numerical results are displayed, providing strong empirical evidence of the relevance of the depth function we propose here.
\end{abstract}

\section{Introduction}

Since its introduction in \citet{Tukey75}, the concept of statistical depth has become increasingly popular in multivariate data analysis. For a distribution $P$ on $\mathbb{R}^d$ with $d>1$, by transporting the natural order on the real line to $\mathbb{R}^d$, a depth function $D(.,\; P):\mathbb{R}^d\rightarrow \mathbb{R}_+$ provides a center-outward ordering of points in the support of $P$ and can be straightforwardly used to extend the notions of (signed) rank or order statistics to multivariate data, which find numerous applications in Statistics and Machine Learning (e.g. robust inference, hypothesis testing, novelty/anomaly detection), see \citet{Mosler13} for instance. Numerous definitions have been proposed, as alternatives to the earliest proposal, the \textit{halfspace} depth introduced in \citet{Tukey75}: among many others, the simplicial depth (see \citet{Liu}), the projection depth (\citet{Liu92}),  the majority depth (\citet{LiuSingh}), the Oja depth (\citet{OJA1983}), the zonoid depth (\citet{koshevoy1997}), the spatial depth (\citet{Chaudhuri} or \citet{Vardi1423}) or the Monge-Kantorovich depth (see \citet{chernozhukov2017}). In order to compare systematically their merits and drawbacks, \citet{ZuoS00a} have devised an axiomatic nomenclature of statistical depths, listing key properties that should be ideally satisfied by a depth function. Roughly, as depth functions serve to define center-outward orderings, if a distribution $P$ on $\mathbb{R}^d$ has a unique center $\theta\in \mathbb{R}^d$ (i.e. a symmetry center in a certain sense), the latter should be the deepest point and the depth should decrease along any fixed ray through it. One also expects that a depth function vanishes at infinity and does not depend on the coordinate system chosen. This latter property is usually formulated as \textit{affine-invariance}. A more formal description of these four properties is given in Section \ref{sec:background}. Beyond the verification of these properties, the pros and cons of any data depth should be considered with regard to the possible existence of algorithms for exact computation in the case of discrete/sampling distributions. In this respect, the extension of Tukey's halfspace depth recently introduced in \citet{IRW} and referred to as the  integrated rank-weighted (IRW) depth offers many advantages. Rather than computing, for any point $x$ in $\mathbb{R}^d$, the minimum of the mass $P(\mathcal{H})$ taken over all closed
half-spaces $\mathcal{H}=\{ x'\in \mathbb{R}^d:\;\; \langle x'-x, u_{\mathcal{H}} \rangle\leq 0 \}$ with unit normal vector $u_{\mathcal{H}}\in \mathbb{R}^d$ and containing $x$, it is proposed to replace the infimum by the integral taken w.r.t. all possible directions $u_{\mathcal{H}}$ uniformly (w.r.t. the uniform distribution on the unit sphere),
following in the footsteps of the general \textit{integrated dual depth} approach developed in \citet{cuevas2009}. For a discrete/empirical distribution, the depth thus constructed admits a weighted average representation and can be easily approximated by means of Monte-Carlo methods in contrast to many other depth functions, whose values are defined as solutions of optimization problems, possibly very complex in high dimension. Beyond these computational aspects, it is shown in \citet{IRW} that the IRW data depth satisfies several desirable properties, see Theorem 2 therein. Unfortunately, it does not fulfill the \textit{affine-invariance} property, as the values taken by the IRW depth may possibly highly depend on the coordinate system chosen to represent the statistical information available, as shown by an illustrative example in the Appendix section. It is the main purpose of this paper to overcome the lack of affine invariance of the IRW depth by proposing a modified version of it, named AI-IRW. It consists in the IRW depth of the (supposedly square integrable) random vector $X$ with distribution $P$ under study expressed in an orthogonal coordinate system such that its components are linearly uncorrelated, i.e. of the random vector whose components are the the principal components of $X$ obtained by eigenvalue decomposition of its covariance matrix $\Sigma$ (Principal Component Analysis). Under the assumption that $\Sigma$ is definite positive (otherwise, the methodology promoted should be naturally applied after a dimensionality reduction step,  i.e. applied to an appropriate orthogonal projection of the original random vector $X$), the \textit{affine-invariant} version of the IRW depth (the AI-IRW depth in abbreviated form) of $X$ is the IRW depth of $\Sigma^{-1/2}X$, denoting by $\Sigma^{-1/2}$ the inverse of the matrix $\Sigma^{1/2}$, the square root of the symmetric positive definite matrix $\Sigma$. In this article, we show that the AI-IRW depth inherits all the properties and computational advantages of the IRW depth and satisfies the \textit{affine-invariance} property in addition. Because its statistical counterpart based on a sample composed of independent copies of the random variable $X$ is a complex functional of the data, involving the square root of the empirical precision matrix, a finite-sample analysis is carried out here. Precisely, a concentration result for the sampling version of the AI-IRW depth is established. Beyond this theoretical analysis, the relevance of the AI-IRW depth notion is also supported by experimental results, showing its advantages over the IRW depth and other depth proposals standing as natural competitors when applied to various statistical tasks such as anomaly detection.

\par The article is structured as follows. In Section \ref{sec:background}, the concept of data depth is briefly reviewed, together with illustrating examples, the integrated rank-weighted depth in particular, and the axiomatic approach developed by \citet{ZuoS00a}. In Section \ref{sec:def}, the AI-IRW depth is introduced, its properties are studied and approximation/estimation issues are discussed at length. The accuracy of the empirical version is investigated in Section \ref{sec:main} from a nonasymptotic perspective. Section \ref{sec:num} describes experimental results illustrating empirically the advantages of the AI-IRW depth. Finally, some concluding remarks are collected in Section \ref{sec:concl}.  Additional technical details and numerical results are deferred to the Appendix section. 

\section{Background and Motivations}\label{sec:background}
The concept of depth function is motivated by the desire to extend the very useful notions of order and (signed) rank statistics in univariate statistical analysis to multivariate situations by means of depth-induced contours. Indeed, such statistics serve to perform a wide variety of tasks, ranging from robust statistical inference to efficient statistical hypothesis testing for instance.
The earliest proposal is the halfspace depth developed in  \citet{Tukey75}, whose popularity arises in particular from its strong connection with the notion of distribution function in the univariate context. Indeed, for any probability measure $P_1$ on $\mathbb{R}$, constructed as a median-oriented distribution function,
the univariate halfspace depth is given by: $\forall t\in \mathbb{R}$,
\begin{equation*}\label{univariate}
 D_{\mathrm{H,1}}(t,P_1) = \min \left\{ P_1\left([-\infty, t] \right), P_1\left((t,+\infty ]\right)\right\}.
\end{equation*}
Considering a multivariate r.v. $X$ with probability distribution $P$ on $\mathbb{R}^d$ with $d>1$, its halfspace depth at $x\in \mathbb{R}^d$ is then defined as the infimum of the probability mass taken over all possible closed halfspaces containing $x$:

\vspace{-0.8cm}

\begin{align}\label{multivariate}
D_{\mathrm{H}}(x,P) = \underset{u\in \mathbb{S}^{d-1}}{\inf} \mathbb{P} \left( \langle u,X\rangle \; \leq \; \langle u,x\rangle \right),
\end{align}

\vspace{-0.3cm}
\noindent denoting by $\langle \cdot, \cdot \rangle$ and $\vert\vert \cdot \vert\vert$ the usual Euclidean inner product and norm on $\mathbb{R}^d$, by $\mathbb{S}^{d-1}=\{ z\in \mathbb{R}^d:\; \vert\vert z \vert\vert=+1 \}$ the unit sphere of $\mathbb{R}^d$ w.r.t. the Euclidean norm.
The halfspace depth \eqref{multivariate}, probably because of some of its appealing properties (it is quasi-concave, upper semi-continuous), is undeniably the most documented notion of depth function in the statistical literature.  It has been proved to fully characterize discrete/empirical distributions in \citet{StruyfR99,Koshevoy02}. Asymptotic guarantees (consistency, asymptotic normality) for its sampling version based on independent copies $X_1,\; \ldots,\; X_n$ of the generic r.v. $X$ (obtained by replacing $P$ in \eqref{multivariate} with the empirical distribution $\widehat{P}=(1/n)\sum_{i=1}^n\delta_{X_i}$, where $\delta_x$ means the Dirac mass at any point $x$) are established in e.g. \citet{RousseeuwS98,Donoho82,ZuoS00a}. Multivariate location estimators based on it have been investigated in \citet{DonohoG92} and it has been shown to possess attractive robustness properties. For instance, the asymptotic breakdown point of the Tukey median, i.e. the barycenter of the deepest locations in the sense of \eqref{multivariate}, is equal to $1/3$ for absolutely continuous centrosymmetric distributions, see \citet{DonohoG92}. Computational issues have also been extensively studied, see \citet{LiuZ14a} or \citet{LiuMM18} for instance. However, as recalled in the Introduction section, many other notions of depth have been proposed these last decades, far too numerous to be listed in an exhaustive manner here, refer to \citet{Mosler13} for an excellent account of the statistical depth theory. In order to guarantee the  ‘‘center-outward ordering'' interpretation of a depth function $D(., P): \mathbb{R}^d \rightarrow \mathbb{R}_+$ of a probability distribution $P$ on $\mathbb{R}^d$, four key properties have been listed by \citet{ZuoS00a}, see also \citet{Dyckerhoff04} and \citet{Mosler13} for a different formulation of the latter. They are recalled below.

\begin{itemize}
\item[$\mathbf{D}_1$] {\sc (Affine invariance)} Denoting by $P_X$ the distribution of any r.v. $X$ taking its values in $\mathbb{R}^d$,  we have
\begin{equation*}\label{eq:AI}
\forall x\in \mathbb{R}^d,\;\;  D(Ax+b, P_{AX+b})=D(x, P_X),
\end{equation*}
for any $d$-dimensional r.v. $X$, any $d\times d$ nonsingular matrix $A$ with real entries and any vector $b$ in $\mathbb{R}^d$.
\item[$\mathbf{D}_2$] {\sc (Maximality at center)} For any probability distribution $P$ on $\mathbb{R}^d$ that possesses a symmetry center $x_P$ (in a sense to be specified), the depth function $D(., P)$ takes its maximum value at it: 
\begin{equation*}\label{eq:MC}
D(x_P,\; P)=\sup_{x\in \mathbb{R}^d}D(x, P).
\end{equation*}
\item[$\mathbf{D}_3$] {\sc (Monotonicity relative to deepest point)} For any probability distribution $P$ on $\mathbb{R}^d$ with deepest point $x_P$, the depth at any point $x$ in $\mathbb{R}^d$ decreases as one moves away from $x_P$ along any ray passing through it: \begin{equation*}\label{eq:MoDP}
\forall \alpha \in [0,1],\;\; D(x_P, P)\geq D(x_P+\alpha(x-x_P), P).
\end{equation*}
\item[$\mathbf{D}_4$] {\sc (Vanishing at infinity)} For any probability distribution $P$ on $\mathbb{R}^d$, the depth function $D$ vanishes at infinity:
\begin{equation*}\label{eq:Van}
D(x, P)\rightarrow 0, \text{ as } \vert\vert x\vert\vert\rightarrow \infty.
\end{equation*}
\end{itemize}

It is worth mentioning that the most general notion of symmetry when analyzing data depth is the halfspace symmetry and will be the one we rely on in the paper. Precisely, the probability distribution $P$ is halfspace symmetric in $x_P$ if $P(\mathcal{H}_{x_P})\geq 1/2$ for all closed halfspaces $\mathcal{H}_{x_P}$ passing through $x_P$.

Various works have examined which of the properties, among those listed above, are satisfied by specific notions of depth introduced in the literature, see \citet{ZuoS00a}. Some of them are constructed as an infimum over unit-sphere projections of a univariate non parametric statistic such as the projection depth proposed by \citet{Liu92} or those introduced in \citet{zhang2003} or \citet{Zuo03}. From a practical perspective, computing these projection-based depths involves the use of tools such as manifold optimization algorithms, facing various numerical difficulties as the dimension $d$ increases, see \citet{dyckerhoff2020approximate}. In addition, the halfspace depth suffers from two major problems: (i) for each data point, taking the direction achieving the minimum to assign a score to it possibly creates a significant sensitivity to noisy directions  (ii) the null score assigned to each new data point outside of the convex hull of the support of the distribution $P$ makes the score of such points indistinguishable.  A remedy based on Extreme Value Theory has been proposed in \citet{EinmahlLL15}, which consists in smoothing the halfspace depth beyond the convex hull of the data. However, this variant relies on rather rigid parametric assumptions, is only approximately affine invariant and confronted with the limitation aforementioned regarding the non-smoothed part of the data. Recently, alternative depth functions have been proposed, obtained by replacing the infimum over all possible directions by an integral, see \citet{cuevas2009}. In \citet{IRW}, a new data depth, referred to as the Integral Rank-Weighted (IRW) depth, is defined by substituting an integral over the sphere $\mathbb{S}^{d-1}$ for the infimum in \eqref{multivariate}. Here and throughout, the indicator function of any event $\mathcal{E}$ is denoted by $\mathbb{I}\{ \mathcal{E} \}$, the spherical probability measure on $\mathbb{S}^{d-1}$ by $\omega_{d-1}$.

\begin{definition}[\citet{IRW}] The Integrated Rank-Weighted  (IRW) depth of $x\in \mathbb{R}^d$ relative to a probability distribution $P$ on $\mathbb{R}^d$ is given by: 
\vspace{-0.1cm}
\begin{align}\label{eq:IRW}
D_{\text{IRW}}(x,P) &= \int_{\mathbb{S}^{d-1}} D_{\mathrm{H,1}}(\langle u,\;  x\rangle , P_u)\;\omega_{d-1}(du)
\\&= \mathbb{E}\left[D_{\mathrm{H,1}}(\langle U,\;  x\rangle, P_U) \right] \nonumber,
\end{align}
where $P_u$ is the pushforward distribution of $P$ defined by the projection $x\in \mathbb{R}^d\mapsto \langle u,x\rangle$ and $U$ is a r.v. uniformly distributed on the hypersphere $\mathbb{S}^{d-1}$.
\end{definition}

As explained at length in \citet{IRW}, the name of the data depth \eqref{eq:IRW} originates from the fact that it can be represented as a weighted average of a finite set of normalized center-outward ranks. It has many advantages over the original halfspace depth \eqref{multivariate}. First, by construction it is \textit{robust} to noisy directions and \textit{sensitive} to new data point outside of the convex hull of the training data set both at the same time, fixing then the two problems mentioned above. Moreover, concerning numerical feasibility, the computation of the IRW depth does not require to implement any  manifold optimization algorithm and can be approximately made by means of basic Monte Carlo techniques, providing in addition confidence intervals as a by-product, see Remark \ref{rk:MC} below. Its contours $\{ D_{\text{IRW}}(x,P)=\alpha\}$, $ \alpha\in [0,1]$, also exhibits a higher degree of smoothness in general (the depth function \eqref{eq:IRW} is continuous at any point $x\in \mathbb{R}^d$ that is not an atom for $P$, cf. Proposition 1 in \citet{IRW}) and properties $\mathbf{D}_2$, $\mathbf{D}_3$ and $\mathbf{D}_4$ have been proved to be satisfied by \eqref{eq:IRW} under mild assumptions, see Theorem 2 in \citet{IRW}.

\begin{remark}\label{rk:MC} ({\sc Monte Carlo approximation}) Recall that a r.v. uniformly distributed on the hypersphere $\mathbb{S}^{d-1}$ can be generated from a $d$-dimensional centered Gaussian
random vector $W$ with the identity $\mathcal{I}_d$ as covariance matrix: if $W\sim \mathcal{N}(0, \mathcal{I}_d)$, then $W/\vert\vert W\vert\vert \sim \omega_{d-1}$, see \citet{Krantz}. Hence, a basic Monte-Carlo method to approximate \eqref{eq:IRW} would consist in generating $m\geq 1$ independent realizations $W_1, \ldots, W_m$ of $\mathcal{N}(0, \mathcal{I}_d)$ and compute
\begin{equation}\label{eq:MC}
\frac{1}{m}\sum_{j=1}^m D_{\mathrm{H,1}}(\langle W_j/\vert\vert W_j\vert\vert,\;  x\rangle , P_{W_j/\vert\vert W_j\vert\vert}),
\end{equation}
refer to e.g. \citet{Kalos} for an account of Monte Carlo integration methods.
\end{remark}

 However,  it does not satisfy the key property $\mathbf{D}_1$ (affine-invariance) in general as illustrated in the next section (see also Section ~\ref{exampleAI} and Section ~\ref{figAI} of the Appendix for an analytical example and for an additional numerical illustration, respectively). 
 The fact that it is affected by non-uniform scaling is problematic in practice (regarding its interpretability in particular or its use for anomaly detection tasks for instance, see Section \ref{sec:num}) and is the main flaw of this approach, as pointed out in \citet{cuevas2009,IRW}.


\section{Affine-Invariant IRW Depth - Definition and Properties}\label{sec:def}

Here we propose to modify the depth function \eqref{eq:IRW}  in order to ensure that property $\mathbf{D}_1$ is always satisfied when the random vector $X$ with distribution $P$ under study is assumed to be square integrable with positive definite covariance matrix $\Sigma$. Precisely, rather than taking the expectation w.r.t. a random direction $U$ uniformly distributed on $\mathbb{S}^{d-1}$ (i.e. integrating over all possible directions $u\in \mathbb{S}^{d-1}$), one considers the random projections defined by the eigenfunctions of the matrix $\Sigma$, i.e. the principal components of the r.v. $X$. In other words, the expectation is taken w.r.t. the distribution of the random vector $V=\Sigma^{{\scriptscriptstyle - \top/2}} U/\vert\vert \Sigma^{{\scriptscriptstyle - \top/2}} U \vert\vert$ valued in $\mathbb{S}^{d-1}$, yielding the definition below.


\begin{definition} [{\sc Affine-invariant IRW depth}] 
The Affine-Invariant Integrated Rank-Weighted  (AI-IRW) depth relative to a square integrable random vector $X$ with probability distribution $P$ on $\mathbb{R}^d$ and positive definite covariance matrix $\Sigma$ is given by:

\begin{equation}\label{eq:AIIRW}
\forall x\in \mathbb{R}^d,\;\; D_{\scriptscriptstyle \text{AI-IRW}}(x, P)=\mathbb{E}\left[D_{\mathrm{H,1}}(\langle V,\;  x\rangle, P_{V})\right],
\end{equation}

where $V=\Sigma^{{\scriptscriptstyle - \top/2}} U/\vert\vert \Sigma^{{\scriptscriptstyle - \top/2}} U \vert\vert$ and $U$ is uniformly distributed on the hypersphere $\mathbb{S}^{d-1}$. 

\end{definition}

Of course, in the case where the covariance matrix $\Sigma$ of the supposedly square integrable r.v. $X$ is not invertible, the AI-IRW depth notion should be applied to an orthogonal projection, after an appropriate dimensionality reduction step. From a computational perspective, The AI-IRW depth can be approximated by Monte Carlo methods in the same way as \eqref{eq:IRW}, see Remark \ref{rk:MC}.
As revealed by the proposition stated below, the depth function \eqref{eq:AIIRW} inherits all the properties of \eqref{eq:IRW} under similar assumptions and is remarkably invariant under any affine transformation in addition.

\begin{figure}[!h]
    \centering
    \begin{tabular}{cc}
    \includegraphics[scale=0.14, trim=2cm 0 0 0]{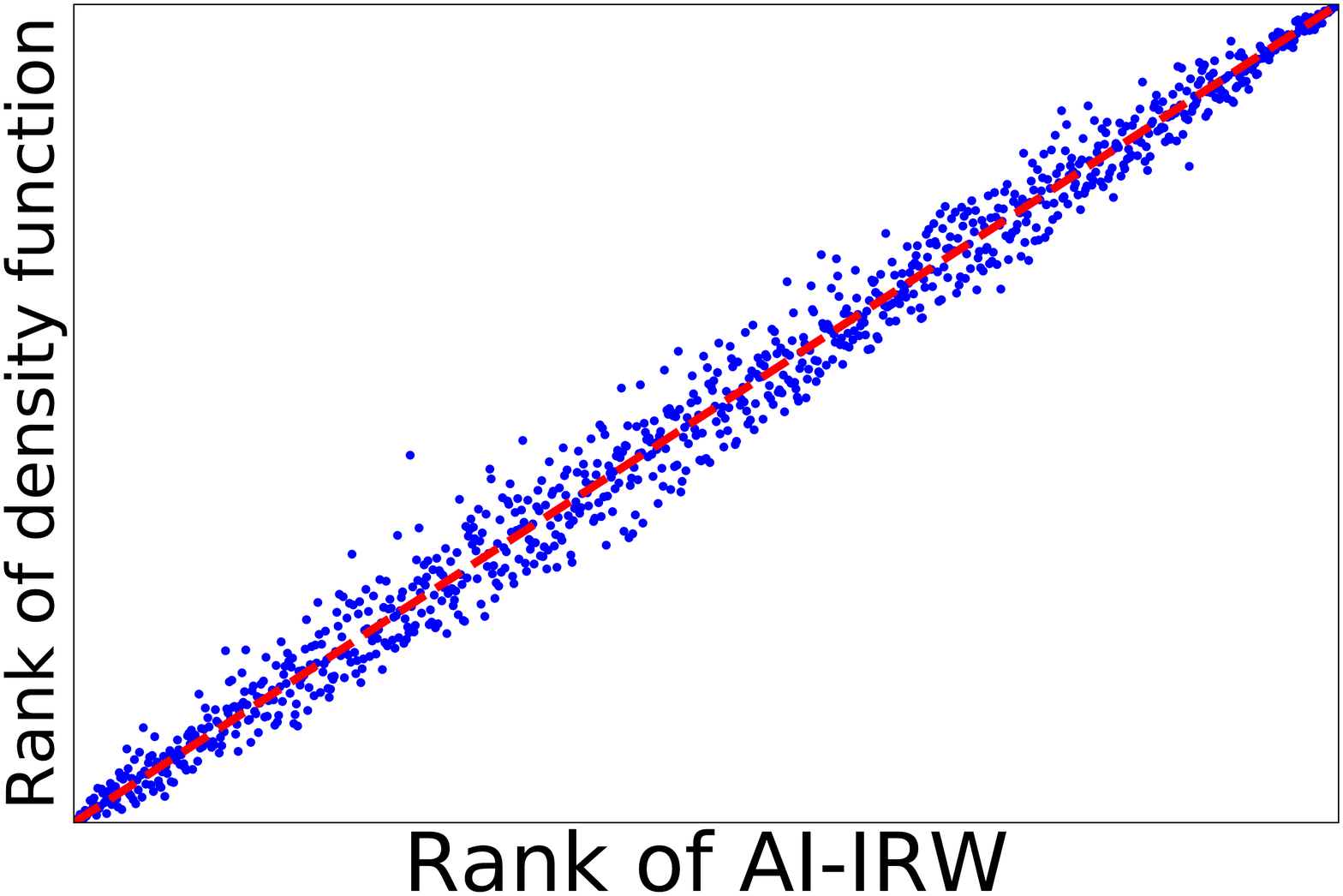}&
    \includegraphics[scale=0.14, trim=2.7cm 0 0 0]{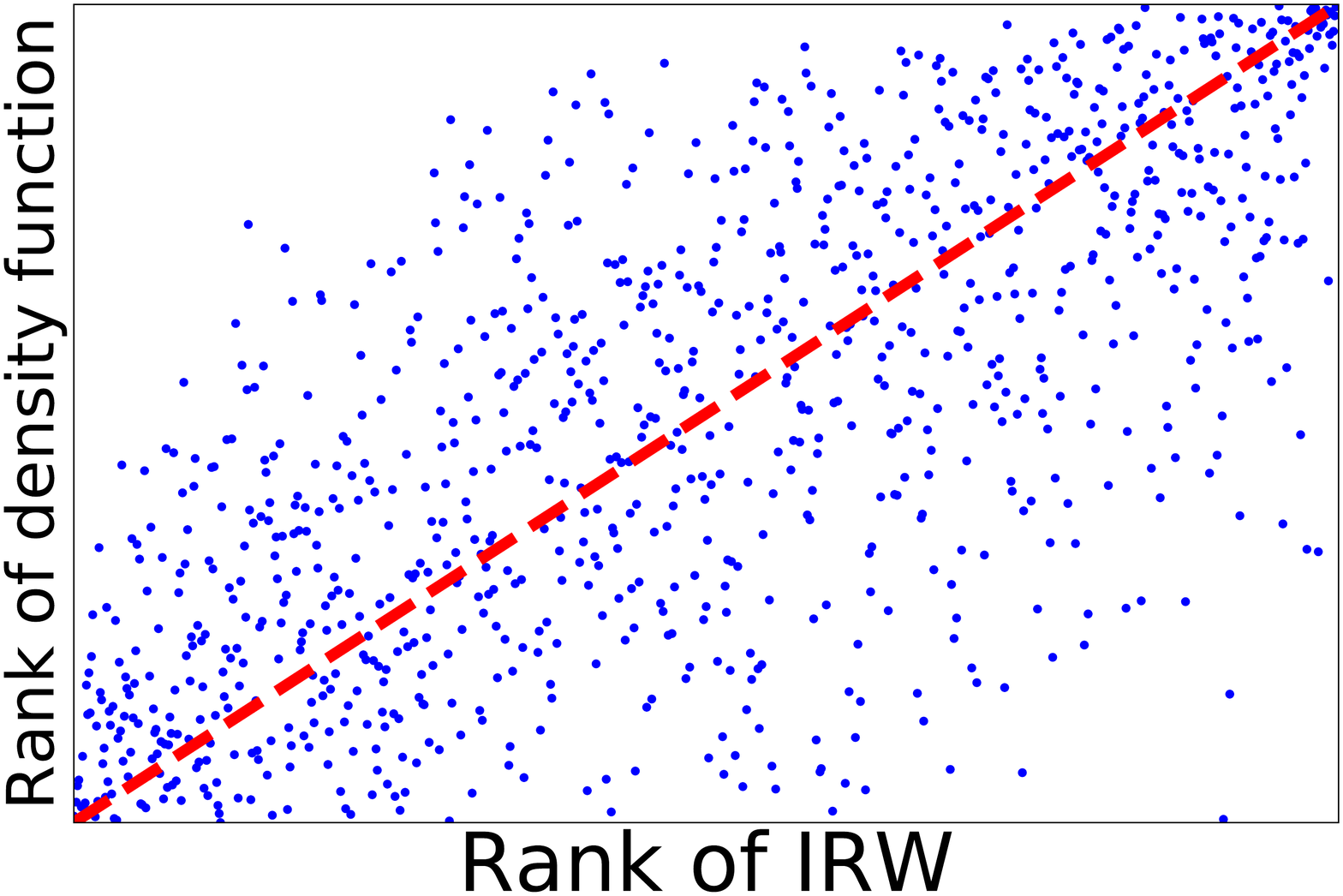}
    \end{tabular}
    \caption{Rank-Rank plots comparing the ranks of $1000$ points sampled from a $10$-d (anisotropic) Gaussian distribution with covariance matrix drawn at random from a Wishart distribution (with parameters ($d,\mathcal{I}_d$)) induced by the empirical depth (AI-IRW on the left, IRW on the right) and those induced by the Gaussian density. }
    \label{fig:exemple}
\end{figure}

\begin{proposition} \label{proposition} The assertions below hold true for any probability distribution $P$ of a square integrable r.v. $X$ valued in $\mathbb{R}^d$ with positive definite covariance matrix.
\begin{itemize}
\item[$(i)$] The AI-IRW depth  satisfies the properties $\mathbf{D}_1$ and $\mathbf{D}_4$. In addition, $\mathbf{D}_2$ and $\mathbf{D}_3$ hold for halfspace symmetric distributions.
\item[$(ii)$] The AI-IRW  depth function is continuous at each point $x$ that is not an atom for $P$.
\end{itemize}
\end{proposition}
The proof is detailed in Section \ref{aiirw:proof:prop} of the Appendix.  It is known that for elliptical distributions, affine invariant data depth level sets are concentric ellipsoids with the same center and orientation as the density level sets \citep{LiuSingh}. Therefore, the ordering returned by affine-invariant data depths should be equal to that of the density function. Thus, in order to highlight the discrepancy between AI-IRW and IRW w.r.t. affine-invariance, we propose to compare the ordering returned by AI-IRW and IRW to that of the density function on the Gaussian distribution which belongs to the family of elliptical distributions. As illustrated by the Rank-Rank plots in Figure~\ref{fig:exemple}, the ordering defined by the (empirical) AI-IRW depth is generally much closer to that induced by the underlying density than the order defined by the original (IRW depth) version.

In practice, the distribution $P$ is generally unknown as well as the covariance matrix $\Sigma$ and only a sample $\mathcal{D}_n=\{X_1,\; \ldots,\; X_n\}$ composed of $n\geq 1$ independent realizations of the distribution $P$ is available. A statistical counterpart of the AI-IRW depth can be obtained by replacing $P$ with the empirical measure $\widehat{P}=(1/n)\sum_{i=1}^{n} \delta_{X_i}$ and $\Sigma^{{\scriptscriptstyle - \top/2}}$ with an estimator $\widehat{\Sigma}^{{\scriptscriptstyle - \top/2}}$ based on $\mathcal{D}_n$ and plugging them next into formula \eqref{eq:AIIRW} when $\widehat{\Sigma}$ is invertible, yielding: $\forall x\in \mathbb{R}^d$,
\begin{equation}\label{eq:AIIRW_emp}
\widehat{D}_{\scriptscriptstyle \text{AI-IRW}}(x)=\mathbb{E}\left[D_{\mathrm{H,1}}(\langle \widehat{V},\;  x\rangle, \widehat{P}_{\widehat{V}} ) \mid \mathcal{D}_n  \right],
\end{equation}
where $\widehat{V}=\widehat{\Sigma}^{{\scriptscriptstyle - \top/2}} U/\vert\vert \widehat{\Sigma}^{{\scriptscriptstyle - \top/2}} U \vert\vert$  and $U$ is a r.v. uniformly distributed on $\mathbb{S}^{d-1}$ independent from the $X_i$'s. 
From a practical perspective, the (conditional) expectation \eqref{eq:AIIRW_emp} can also be approximated by means of a basic Monte Carlo scheme, generating $m\geq 1$ i.i.d. random directions $U_1,\ldots, U_m$,  copies of the generic r.v. $U$ and independent from the original data $\mathcal{D}_n$: $\forall x\in \mathbb{R}^d$,

\begin{align} \label{empiricalformulae}
&\widetilde{D}_{\scriptscriptstyle \text{AI-IRW}}^{\text{MC}}(x) = \nonumber \\& \frac{1}{ m} \; \sum_{j=1}^{m} \min \left\{\widehat{F}_{\widehat{V}_j}\left(\langle \widehat{V}_j ,\;  x  \rangle \right), 1-\widehat{F}_{\widehat{V}_j}\left(\langle \widehat{V}_j ,\;  x  \rangle \right) \right\}, 
\end{align}

\noindent where, for all $j\in \{1,\; \ldots,\; m\}$ and $t\in \mathbb{R}$, we set $$\widehat{V}_j=\widehat{\Sigma}^{{\scriptscriptstyle - \top/2}} U_j/\vert\vert \widehat{\Sigma}^{{\scriptscriptstyle - \top/2}} U_j \vert\vert, $$ and $$  \widehat{F}_{\widehat{V}_j}(t) = \frac{1}{n}\sum_{i=1}^n \mathbb{I}_{\left\{ \langle \widehat{V}_j, X_i \rangle \leq t \right\}},$$
where $\mathbb{I}_{\mathcal{E}}$ means the indicator function of any event $\mathcal{E}$. Putting aside the issue of estimating $\Sigma^{{\scriptscriptstyle - \top/2}}$ (discussed below), attention should be paid to the fact that the approximate sample version \eqref{empiricalformulae} is very easy to compute (refer to Section~\ref{algoapprox} in the Appendix for further details) and involves no optimization procedure, in contrast to many other notions of depth function.
\medskip

\noindent {\bf On estimating the square root of the precision matrix.} Consider the $d\times n$ matrix $\mathbf{X}_n=(X_1, \ldots, X_n)$ with the $X_i$'s as columns. The simplest way of building an estimate $\widehat{\Sigma}^{-\top /2}$ consists in computing the empirical version of the covariance matrix $\widehat{\Sigma}=(1/n)\mathbf{X}_n\mathbf{X}_n^\top$, which is a natural and nearly unbiased estimator, and inverting next its square root, when the latter is positive definite (which happens with overwhelming probability). 
For simplicity, this is the estimation we consider in the finite-sample study presented in the next section. However, alternative techniques can be used, yielding possibly more efficient estimators under specific assumptions, in high-dimension especially. Shrinkage procedures for covariance estimation under sparsity conditions have been investigated in e.g. \citet{ledoit,chen2010,schafer}, while a lasso method for direct estimation of the precision matrix, avoiding matrix inversion, is proposed in \citet{friedman}. Robust covariance estimation techniques, tailored to situations where the data are possibly contaminated or heavy-tailed, have also been documented in the literature, see e.g. \citet{MCD}  and \citet{fastMCD}. Classically, from a symmetric definite positive estimator of the covariance matrix, one can easily build an estimator of the square root of the precision matrix by inverting a triangular/diagonal matrix.
Due to the presence of $\widehat{\Sigma}^{-\top /2}$ in \eqref{eq:AIIRW_emp} (respectively, in \eqref{empiricalformulae}), it is far from straightforward to assess the accuracy of the estimators of the AI-IRW depth proposed above. It is the purpose of the next section to study the uniform deviations between \eqref{eq:AIIRW} and its empirical versions from a nonasymptotic perspective.

\section{Finite-Sample Analysis - Concentration Bounds}\label{sec:main}

We now investigate the accuracy of the statistical version, as well as that of its Monte Carlo approximation, of the  AI-IRW depth function introduced in the previous section in a nonasymptotic fashion. Precisely, we establish a concentration bound for the maximal deviations between the true and estimated AI-IRW depth functions. We assume here that the estimator of the square root of the precision matrix is given by the inverse of the square root of the empirical covariance, when the latter is definite positive (which happens with overwhelming probability), and by that of any definite positive regularized version (e.g. Tikhonov) of the latter otherwise.
The subsequent analysis requires additional hypotheses, listed below.
The first assumption, classical when estimating the precision matrix (see e.g. \citet{cai2013} or \citet{fan2015overview}), stipulates that the eigenvalues $\sigma_1, \ldots, \sigma_d$ of the covariance matrix $\Sigma$ of the square integrable random vector $X$ considered are bounded away from zero.

\begin{assumption}\label{as:1} There exists $\varepsilon>0$ such that: $\forall  k\in\{1,\; \ldots,\; d\}$, $\varepsilon\leq \sigma_k$. 
\end{assumption}
The second assumption is technical, see \citet{daviskahan}. It stipulates that $\Sigma$'s eigenvalues are all of multiplicity $1$ and that $\Sigma$'s minimum eigengap is bounded away from zero.

\begin{assumption}\label{as:1_bis} There exists $\gamma>0$ such that: $\forall  k\in\{1,\; \ldots,\; d-1\}$, $\gamma\leq \sigma_{(k)}-\sigma_{(k+1)}$, where $\sigma_{(1)}> \ldots > \sigma_{(d)}$ are $\Sigma$'s eigenvalues sorted by decreasing order of magnitude. 
\end{assumption}
We point out that, just like when $\Sigma$ is not invertible, one always may bring back the analysis to a situation where Assumption \ref{as:1_bis} is fulfilled by means of a preliminary dimensionality reduction step. Notice incidentally that, when $\Sigma=\sigma \mathcal{I}_d$, with $\sigma>0$, the AI-IRW reduces to IRW. The other assumptions correspond to smoothness conditions of Lipschitz type for the function $\phi: (u,x)\in \mathbb{S}^{d-1}\times \mathbb{R}^d \mapsto \mathbb{P}\left\{\langle u,X\rangle \leq \langle u,x \rangle \right\}$.

\begin{assumption} \label{as:2} {\sc (Uniform Lipschitz condition in projection)} For all $(x, y)\in \mathbb{R}^d\times \mathbb{R}^d$, there exists $L_{p}<+\infty$ such that 
\begin{align*}
\underset{u \in \mathbb{S}^{d-1}}{\sup } \;  |\phi(u,x) - \phi(u,y)| \leq L_p ||x-y||.
\end{align*}
\end{assumption}

\begin{assumption}\label{as:3} {\sc (Uniform radial Lipschitz condition)} For all $(u, v)\in \mathbb{S}^{d-1} \times \mathbb{S}^{d-1} $, there exists $L_{R}<+\infty$ such that 
\begin{align*}
\underset{x \in \mathbb{R}^d}{\sup } \; |\phi(u,x) - \phi(v,x)| \leq L_R ||u-v||.
\end{align*}
\end{assumption}

\vspace{-0.35cm}
Notice that the same assumptions are involved in the non-asymptotic rate bound analysis carried out for the halfspace depth estimator in \citet{burr} and are used to establish limit results related to its approximation in \citet{nagyuniformrates}. The Lipschitz conditions are satisfied by a large class of probability distributions, for which Lipschitz constants $L_R$ and $L_p$ can be both explicitly derived. For instance, if the distribution $P$ of $X$ has compact support included in the ball $\mathcal{B}(0,r)=\{x\in \mathbb{R}^d:\; \vert\vert x \vert\vert\leq r \}$ (relative to the Euclidean norm $\vert\vert \cdot\vert\vert$) with $r>0$ and is absolutely continuous w.r.t. the Lebesgue mesure with a density bounded by $M>0$, the uniform Lipschitz conditions are then fulfilled with $L_R=M V_{d,r}$ and $L_p= MV_{d-1,r}$, where $V_{d,r}=\pi^{d/2}r^d/\Gamma(d/2+1)$ is the volume of the ball $\mathcal{B}(0,r)$ and $z\geq 0\mapsto \Gamma(z)=\int_0^{\infty}t^{z-1}e^{-t}dt$ means the Gamma function,  refer to the Appendix section for further details (see Lemma \ref{lemma:argball} and \ref{lemma:argball2} therein) and to \citet{burr} for additional examples. In contrast, a necessary condition for Assumption \ref{as:2} to be satisfied is the absolute continuity of the measure $P$ w.r.t. the Lebesgue measure, see Section 4 in \citet{nagyuniformrates}. The bounds stated in the theorem below reveal the accuracy of the statistical estimates \eqref{eq:AIIRW_emp} and \eqref{empiricalformulae} and highlight their behavior through explicit constants.




\begin{theorem}\label{theorem} Suppose that the distribution $P$ of the r.v. $X$ is $\tau$ sub-Gaussian and satisfies Assumptions \ref{as:1}, \ref{as:1_bis}, \ref{as:2} and \ref{as:3}. The following assertions hold true.
\begin{itemize}
\item[(i)] For any $\delta\in \left(\max\{\Theta, 12.9^d\} ~ e^{-\frac{n}{2}\min \left\{\alpha,\alpha^2, \alpha \Delta / 8 \right\}},1\right)$, we have with probability at least $1- \delta$:

\begin{align*}
&\underset{x \in \mathbb{R}^d}{\sup }\;   \Big \lvert \widehat{D}_{\scriptscriptstyle \text{AI-IRW}}(x) -    D_{\scriptscriptstyle \text{AI-IRW}}(x, P) \Big \rvert \\&\leq  \Delta(L_R,d,\gamma, \varepsilon, \tau)~ \underset{s=1,2}{\max} \left( \dfrac{d+\log(2/\delta)}{n} \right)^{1/s} \\& \quad+ \sqrt{\dfrac{8\log(\Theta/\delta)}{n}},
\end{align*}
where $\Delta=512L_R\tau^2 \max\{ 1/\xi ,\; 2\sqrt{2d} /\gamma \}$  with $\xi\in (0,\varepsilon)$, $\alpha(\varepsilon, \tau) = (\varepsilon- \xi)/(32\tau^2)$ and $\Theta=12 (2n)^{d+1}/(d+1)!$.

\item[(ii)] Let $r>0$. For any $\delta\in \left(\max\{\Theta ,\; 12.9^d\}~  e^{-n\min \left\{\alpha,\alpha^2, \alpha \Delta / 8 \right\}},1\right)$, we have with probability at least $1- \delta$:
\begin{align*}
&\underset{x \in \mathcal{B}_r}{\sup } \;  \Big \lvert \widetilde{D}_{\scriptscriptstyle \text{AI-IRW}}^{\text{MC}}(x) -    D_{\scriptscriptstyle \text{AI-IRW}}(x, P) \Big \rvert \\& \leq \sqrt{\frac{128\log(3\Theta/2\delta)}{9n}} + 2\sqrt{\frac{d \log \left(3rm\right) + \log(6 /\delta)}{18m}}
  \\& \quad +\frac{4 L_p }{3m} +  \frac{8\Delta}{3} \underset{s=1,2}{\max}  \left(\dfrac{d+\log(2/\delta)}{n} \right)^{1/s},
\end{align*}
where the constants $\Theta$, $\Delta$, $\alpha$ and the parameter $\xi\in (0,\varepsilon)$ are the same as those involved in $(i)$.
\end{itemize}

\end{theorem}
Due to space limitations, the detailed proof is postponed to the Section \ref{aiirw:proof:theorem} of the Appendix. The upper confidence bound in assertion $(i)$ is decomposed into two terms. The first term, of order $O(n^{-1/2})$, owes its presence to the replacement of $\Sigma^{-1/2}$ by its estimator. The second term, of order $O(\sqrt{\log(n)/n} )$ and exhibiting a sublinear dependence in the dimension $d$ , corresponds to the bound that would be obtained if $\Sigma^{-1/2}$ were known (it is then derived by means of the arguments used to study the concentration properties of the empirical halfspace depth, see chapter 26 in \citet{shorack86}). The upper confidence bound in assertion $(ii)$ differs from that in assertion $(i)$ in two respects. First, the additional terms clearly show the effect of the Monte Carlo approximation, which is negligible when $n>>m$. Second, the maximal deviation is taken over a compact subset of $\mathbb{R}^d$. Furthermore, our theoretical analysis can be easily extended to the deviations of the sample version of IRW by simply omitting the term involving the square root of the precision matrix corresponding to the first term of $(i)$ and the last term of $(ii)$ leading to faster rates (see Section~\ref{irwbound} in the Appendix section).

\noindent \textbf{A limited confidence interval.} The proof of the assertion $(i)$ relies on controlling the deviations between the eigenvectors (resp., the inverses of the square-root eigenvalues) of $\widehat{\Sigma} $ and those of the true covariance matrix. The lower bound of the $\delta$-range results from this control and is not limiting in practice since it decreases exponentially fast when the sample size increases.

\noindent \textbf{About the constants.} Both upper bounds are provided with explicit constants. 
The explicit linear dependence on the dimension $d$ is due to the operator norm that appears in the proof when controlling the eigenvectors of $\widehat{\Sigma} - \Sigma$. It implies an additional square root of $d$ in the constant $\Delta$ following the classical inequality $||A||_{\text{op}}\leq \sqrt{d} ||A||_1$ for any matrix $A\in \mathbb{R}^{d\times d}$ of full rank $d$. However, Lipschitz constants $L_p$ and $L_R$, that are mandatory in order to derive bounds uniformly on $\mathbb{R}^d$ (or $\mathcal{B}_r$), appear to exhibit an implicit dependence on the dimension $d$. Indeed, these constants can be derived for r.v. valued in a compact support with bounded density exhibiting an exponential dependence on $d$.  Unfortunately, this concern cannot be avoided unless removing the supremum involved in $(i)$ and $(ii)$. While the depth value at a single point $x\in \mathbb{R}^d$ is usually of limited importance, t is often more relevant in practice that an ensemble of depth values, i.e. the set $\{D(x,P),\; x\in \mathbb{R}^d \}$, are simultaneously well approximated by their empirical versions for comparison purposes. This implies estimation guarantees for the ranks induced by the depth function when computed on the whole sample $X_1,\ldots, X_n$, on which several applications such as anomaly detection fully rely on.  The eigengap $\gamma$ appears in the denominator due to the use of a variant of the Davis-Kahan theorem \citep{daviskahan}, so as to control the deviations between the eigenvectors of $\widehat{\Sigma}$ and those of $\Sigma$, and can not be avoided. Observe that both upper-bounds explode as $\gamma$ or $\varepsilon$ vanish. These constants, related to the covariance matrix estimation, are often small in practice (see the Appendix where they are computed on the benchmarked datasets used in Section \ref{sec:num}). However, they are often negligible w.r.t. the lipschitz constant in the numerator that is $O(e^d)$ as mentioned above and is thus not limiting. 



\noindent \textbf{On optimality.} In absence of lower bound (and to the best of our knowledge, no such result is documented in the statistical depth literature yet), the optimality of the bounds above cannot be claimed of course.
However, the proof partly consists in bounding the risk of the estimator of the covariance matrix $\Sigma$ and involves the estimation rates given in Lemma~\ref{lemma:covariance:bounds} in the Appendix, which are known to be optimal for sub-Gaussian distributions \citep{vershynin2012}. 
Recall incidentally that has been shown that faster rates for the estimation of the inverse of the covariance matrix can be established under additional sparsity assumptions (see e.g. Theorem 5 in \citet{cai2010}). 
 
 \noindent \textbf{Choosing $m$.} The difficulty of approximating an integral over  $\mathbb{R}^d$ by means of Monte-Carlo techniques grows with $d$. Our theoretical results, such as the upper bound in $(ii)$, shed light on the behaviour of $m$ w.r.t. the dimension $d$. Indeed, focusing on the term $4L_p/(3m)$, $L_p$ can be explicited for density bounded distributions involving the volume of the unit sphere $\mathbb{S}^{d-1}$ that depends exponentially on $d$ (see the paragraph above Theorem~\ref{theorem}). Thus, $m$ should be higher than $O(e^d)$ to yield a good statistical approximation.  However, in practice, since computation times depend on $m$, a trade-off between statistical accuracy (the higher $m$, the better) and computational burden (the higher $m$, the heavier) must be found in practice, see Section~\ref{sec:num}.

\begin{remark}{\sc (Related work)} We point out that nonasymptotic results about the accuracy of sample versions of statistical depths, such as those stated above, are seldom in the literature.
To the best of our knowledge, rate bounds have only been derived in the halfspace depth case before. The first result (see \citet{shorack86} chapter 26), where uniform rates of the sample version are provided,  uses the fact that the set of halfspaces in $\mathbb{R}^d$ is of finite {\sc VC} dimension. Recently, this result has been refined under the Assumptions \ref{as:2} and \ref{as:3} in \citet{burr}. Asymptotic rates of convergence for the Monte Carlo approximation of the halfspace depth, i.e., when the minimum over the unit hypersphere is approximated from a finite number of directions, have been recently established in \citet{nagyuniformrates}. In contrast to the finite-sample framework, uniform asymptotic rates have been proved in several settings. Unfortunately, approximating a minimum over the unit sphere $\mathbb{S}^{d-1}$ using a Monte Carlo scheme is not optimal. Indeed when the distribution is assumed to belong to a bounded subset of $\mathbb{R}^d$ with bounded density, the authors obtain slow rates of order $O((\log(m)/m)^{1/(d-1)})$ suffering from the curse of dimensionality. Futhermore, they show  that obtaining uniform rates of the halfspace depth approximation is not possible in absence of the bounded density assumption, see Section 4.2 in \citet{nagyuniformrates}.
\end{remark}

\begin{figure*}[!h]
\centering
\begin{tabular}{cc}
\includegraphics[scale=0.28, trim= 1cm 0cm 0cm 0cm]{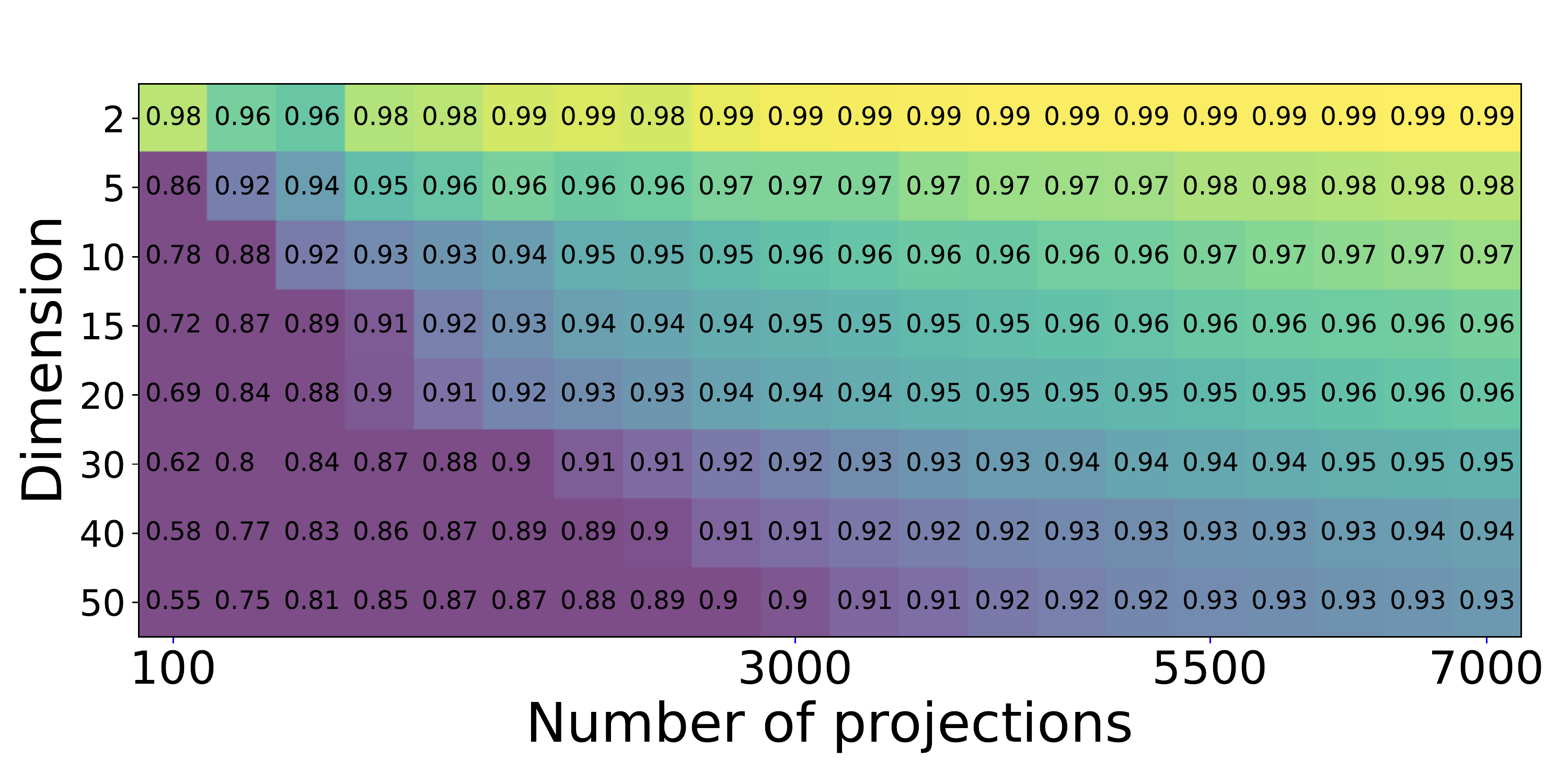}&
\includegraphics[scale=0.28, trim= 1cm 0cm 0cm 0cm]{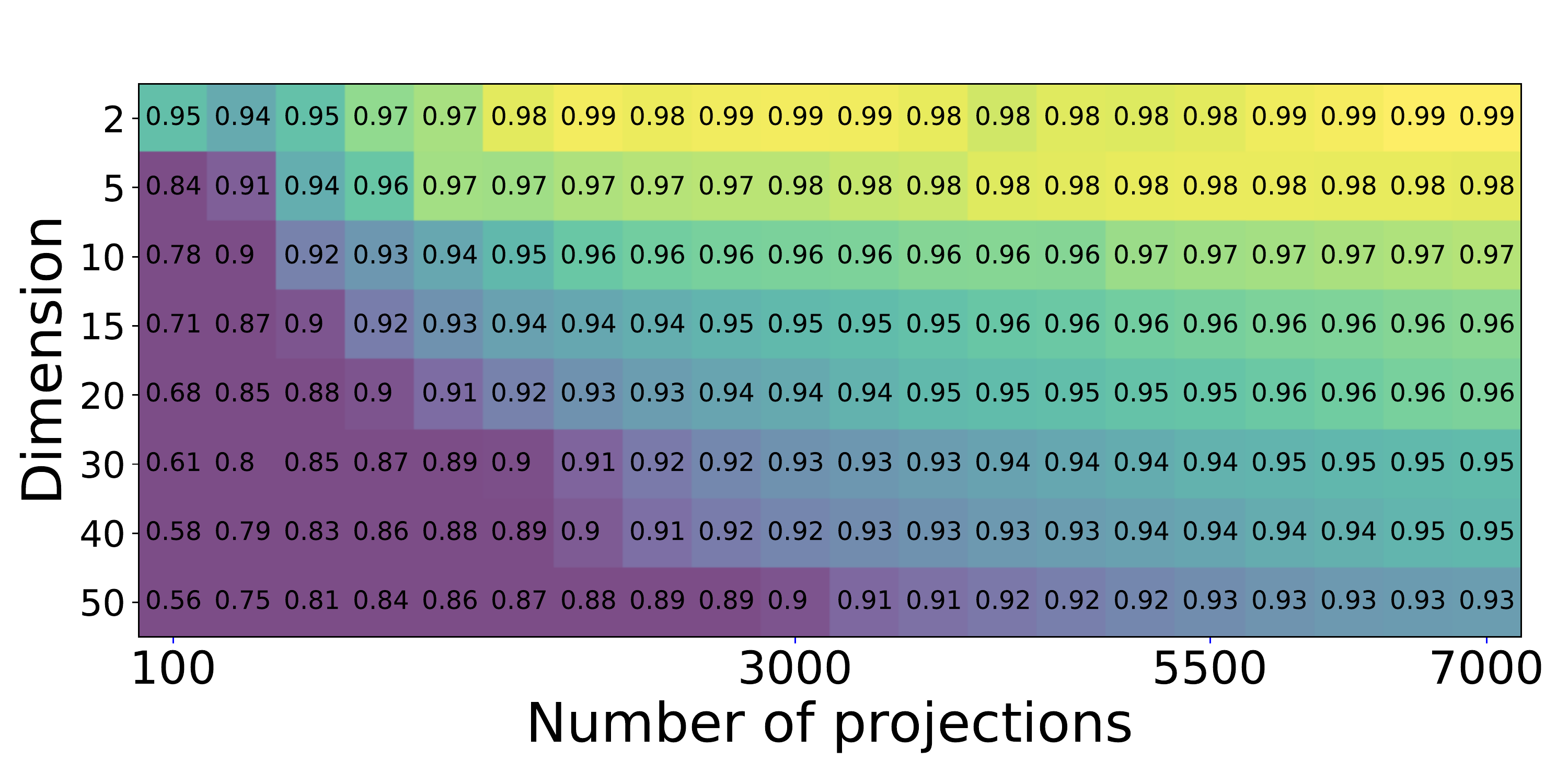}
\end{tabular}
\vspace{-0.5cm}
\caption{Kendall correlation between the approximated ranks of AI-IRW and their true ranks using SC (left) and MCD estimates (right) depending on the number of approximating projections $m$ for a Gaussian distribution.}
\label{fig:projchoice}
\end{figure*}


\section{Numerical Experiments}\label{sec:num}

The advantages of the novel notion of depth introduced in Section \ref{sec:def} are supported by various experimental results in this part.
First, we  explore empirically the behavior of the returned ranks as the number of sampled projections increases. 
Second, the application of the AI-IRW depth to anomaly detection is considered, illustrating clearly the improvement on the performance attained. Due to space limitations, additional experiments are provided in the Appendix. 



\medskip

 \noindent \textbf{On approximating the AI-IRW depth.}
The accuracy of Monte Carlo approximation, depending on the number $m$ of random directions uniformly sampled, is evaluated for the empirical versions of the AI-IRW depth. A robust estimator of the AI-IRW is also introduced using the well-known Minimum Covariance Determinant (MCD)  estimator \citep{MCD} of the covariance matrix. The experiment is based on samples of size $n=1000$ drawn from a centered Gaussian distribution with covariance matrix sampled from a Wishart distribution (with parameters ($d,\mathcal{I}_d)$), where the dimension $d$ varies in the range $\{2,5,10,15,20,30,40,50 \}$. We compute $\widetilde{D}_{\text{AI-IRW}}^{\text{MC}}$ on these samples by varying the number of projections $m$ between $100$ and $7000$. As AI-IRW does not possess any closed-form, we propose to evaluate the quality of the returned ranks considering  $\widetilde{D}_{\text{AI-IRW}}^{\text{MC}}$ computed with $m=200000$ projections as the ‘‘true'' depth. The coherence between ranks is assessed  using the popular Kendall $\tau$ correlation coefficient, see \citet{kendall1938new}. This whole procedure is repeated 10 times and the averaged results are reported in Figure~\ref{fig:projchoice}. As expected, the quality of the approximation increases with $m$ and decreases with $d$. Interestingly, sharp approximations are obtained with far less than $O(e^d)$ projections. Indeed, in the worst case, i.e. when $d=50$, a correlation of $0.93$ is attained for AI-IRW, using both sample covariance (SC) and MCD (with support fraction set to $(n\hspace{-0.08cm}+\hspace{-0.08cm}d\hspace{-0.08cm}+\hspace{-0.08cm}1)/2$) estimators, with only $5500$ directions which is roughly $100\times d$ while $e^{50}\approx 10^{21} $. In low dimension,  few projections are needed to obtain correlation higher than $0.98$. In view of these results and because of the computation time of the approximations (documented in Section~\ref{subsec:computation} of the Appendix due to space constraints), choosing $m=100d$ appears as a good compromise between statistical accuracy and computation time, as done in the next experiment.

\noindent \textbf{Application to anomaly detection.}
To illustrate the performance improvement due to introduction of affine invariance to the IRW, we conduct a comprehensive comparative study of anomaly detection on $10$ widely used data sets in the literature \footnote{\url{http://odds.cs.stonybrook.edu/}}: \textit{Mulcross}, \textit{Shuttle}, \textit{Thyroid}, \textit{Wine}, \textit{Http}, \textit{Smtp}, \textit{Ecoli}, \textit{Breastw}, \textit{Musk} and \textit{Satimage} varying in size and dimension. In this unsupervised setting (we train all methods on unlabeled data), we use labels only to asses the performance of the methods by Area Under the Receiver Operation Characteristic curve (AUROC). We contrast the proposed approach with the non affine-invariant version, the original halfspace depth (T), halfspace mass depth (HM) \citep{HM}, the AutoEncoder (AE) \citep{AE} where the reconstruction error is used as anomaly score and one of the most used multivariate anomaly detection algorithms: Isolation Forest (IF) \citep{LiuTZ08,staerman19}. The performance of these methods being relatively insensitive to their parameters, they are set by default. Based on the previous experiment, AI-IRW, IRW, and halfspace depths are calibrated with $m=100 \times d$. From Table~\ref{table:FM} one observes that AI-IRW uniformly (and significantly in many cases) improves on standard IRW that is rather comparable with Isolation Forest and the halfspace mass depth. Additional information on the data sets as well as the computation time are given in  Section~\ref{add:anom} in the Appendix. 


{\renewcommand{\arraystretch}{1} 
{\setlength{\tabcolsep}{0.13cm}
\begin{table}[h]
\centering
\begin{tabular}{|c||c|c|c|c|c|c|}
 \hline
&AI-IRW&IRW& HM& T& IF & AE\\
 \hline
Ecoli& 0.85& 0.83& \textbf{0.88}& 0.68& 0.77& 0.64\\
 \hline
 Shuttle& 0.99& 0.99& 0.99& 0.86& 0.99& 0.99  \\
 \hline
 Mulcross& \textbf{1} & 0.98& \textbf{1}& 0.87& 0.96 & \textbf{1}\\
 \hline
 Thyroid &  \textbf{0.98} & 0.80& 0.84& 0.92& 0.97 & 0.97\\
 \hline
 Wine& 0.96& 0.96& \textbf{0.99}& 0.71& 0.8 & 0.72\\
 \hline
 Http & \textbf{1} & 0.95& 0.97& 0.99& \textbf{1} &  \textbf{1}  \\
 \hline
 Smtp&  \textbf{0.96} & 0.77 & 0.74 & 0.85 & 0.90 & 0.82 \\
 \hline
 Breastw&0.97& 0.97& \textbf{0.99}&  0.84& \textbf{0.99} & 0.91 \\
 \hline
 Musk& \textbf{1}&  0.84&  0.97& 0.77&\textbf{1} & \textbf{1} \\
 \hline
 Satimage& \textbf{0.99}& 0.96& 0.98& 0.95& \textbf{0.99} & 0.98  \\
 \hline 
\end{tabular}
\caption{AUROCs of benchmarked anomaly detection methods.}
\label{table:FM}
\end{table}
}}

\section{Conclusion}\label{sec:concl}

In this paper, we have introduced a novel notion of statistical depth (AI-IRW), modifying the original Integrated Rank-Weighted (IRW) depth proposal in \citet{IRW}. It has been shown that the AI-IRW depth does not only inherit all the compelling features of the IRW depth, its theoretical properties and its computational advantages (no optimization problem solving is required to compute it), but also fulfills in addition the affine  invariance property, crucial regarding interpretability/reliability issues. The natural idea at work consists in averaging univariate Tukey halfspace depths computed from random projections of the data onto (nearly) uncorrelated lines, defined by the (empirical) covariance structure of the data, rather than projections onto lines fully generated at random. Though the AI-IRW sample version exhibits a complex probabilistic structure, an estimator of the precision matrix being involved in its definition, a nonasymptotic analysis has been carried out here, revealing its good concentration properties around the true AI-IRW depth. The merits of the AI-IRW depth have been illustrated by encouraging numerical experiments, for anomaly detection purpose in particular, offering the perspective of a widespread use for various statistical learning tasks.

\section*{Acknowledgements}

This work has been funded by BPI France in the context of the PSPC Project Expresso (2017-2021).

\bibliography{AIIRW}
\bibliographystyle{icml2022}

\newpage
\appendix
\onecolumn

This Appendix is organized as follows.

\begin{itemize}
\item Useful preliminary results are stated and proved in Appendix A.

\item The proofs of the results stated in the paper are given in Appendix B.

\item The approximation algorithm to compute the AI-IRW depth is described in Appendix C.

\item Additional experiments are displayed in Appendix D.

\end{itemize}
\section{Preliminary Results}\label{preliminary}
First, we recall some lemmas on linear algebra, halfspace depth and covariance matrix estimation, used in the subsequent proofs.


\subsection{Basics of Linear Algebra}
\vspace{0.3cm}
\noindent Here useful results of linear algebra are recalled for clarity.
\begin{lemma}[\citet{Wedin}, Theorem 4.1]\label{wedin}
 Let $A$ and $B$ be two invertible matrices of size $d\times d$ and $||A||_{\text{op}}$ be the operator norm of matrix $A$.  Then it holds:

\begin{equation}
||A^{-1} - B^{-1}||_{\mathrm{op}} \leq ||A^{-1}||_{\mathrm{op}}  || B^{-1}||_{\mathrm{op}}  || A-B||_{\mathrm{op}}.
\end{equation}
\end{lemma}

\begin{lemma}[\citet{vershynin2012}, Lemma 2.2] \label{epsiloncovering}
Let A be a  matrix of size $d \times d$ and $N_{\rho}$ be an $\rho$-net of $\mathbb{S}^{d-1}$. Then it holds:

\begin{align*}
||A||_{\mathrm{op}} \leq \frac{1}{1-2\rho}  \; \underset{v \in N_{\rho}}{\max }  |v^\top Av|.
\end{align*}

\end{lemma}
\begin{lemma} \label{lemma:covariance:decompo}
Let $A_1$ and $A_2$ be two real symmetric and invertible  matrices  of dimension $d\times d$ with $O_1D_1O_1^\top $ and $O_2D_2O_2^\top$ their eigeinvalues decomposition in orthornormal bases. Then it holds:
\begin{align*}
||A_1^{-1/2} - A_2^{-1/2} ||_{\mathrm{op}} &  \leq  ||D_2^{-1/2} ||_{\mathrm{op}} \left( || D_1^{1/2} - D_2^{1/2} ||_{\mathrm{op}}  \; || D_1^{-1/2}||_{\mathrm{op}}  \;  + || O_1 - O_2 ||_{\mathrm{op}}   \right).
\end{align*}
\end{lemma}

\begin{proof}
\begin{align*}
 ||O_1D_1^{-1/2} -O_2D_2^{-1/2} ||_{\mathrm{op}} & \leq ||O_1 ||_{\mathrm{op}}  \; || D_1^{-1/2} - D_2^{-1/2} ||_{\mathrm{op}}  + || O_1 - O_2 ||_{\mathrm{op}}  \; ||D_2^{-1/2} ||_{\mathrm{op}}  \\& \leq || D_1^{-1/2} - D_2^{-1/2} ||_{\mathrm{op}}  + ||D_2^{-1/2} ||_{\mathrm{op}} || O_1- O_2 ||_{\mathrm{op}}  \\& \overset{(i)}{\leq}  ||D_2^{-1/2} ||_{\mathrm{op}} \left( || D_1^{1/2} - D_2^{1/2} ||_{\mathrm{op}}  \; || D_1^{-1/2}||_{\mathrm{op}}  \;  + || O_1 - O_2 ||_{\mathrm{op}}   \right),
\end{align*}
 where ($i$) holds due to Lemma  \ref{wedin}. 
\end{proof}


\subsection{Non-Asymptotic Rates on Halfspace Depth and Sample Covariance Matrix}

We now recall useful results on maximum deviations of the halfspace depth estimator as well as the sample covariance matrix.

\begin{lemma}[\citet{shorack86}, Chapter 26]\label{tukey} Let $P\in \mathcal{P}(\mathbb{R}^d)$. Let $X_1,\ldots,X_n$ a sample from $P$ with empirical measure $\widehat{P}=(1/n)\sum_{i=1}^{n} \delta_{X_i}$. Denote by $F_u$ and $\widehat{F}_u$ the cdf of $P_u$ and $\widehat{P}_u$ respectively. Then, for any $t>0$, it holds:
\begin{align*}
 \mathbb{P} \left( \underset{\substack{x\in \mathbb{R}^d \\ u \in \mathbb{S}^{d-1}}}{\sup } \; \Big \lvert \widehat{F}_{u}(u ^\top x) -F_{u}(u^\top x) \Big \rvert \; > \;t \right)\leq  \frac{6 (2n)^{d+1}}{(d+1)!}\exp(-nt^2/8).
\end{align*}
\end{lemma}

\begin{lemma}[Variant of  \citet{vershynin2012}, Proposition 2.1]\label{lemma:covariance:bounds}
Let $\Sigma$ be the covariance matrix of a $\tau$ sub-Gaussian random variables $X$ that takes its values in $\mathbb{R}^d$. Let $X_1\ldots,X_n$ be a sample from $X$ and denote by $\widehat{\Sigma}= \frac{1}{n} \sum_{i=1}^{n}X_iX_i^\top$ the SC estimator of $\Sigma$. Then it holds:

\begin{align*}
\mathbb{P} \left(||\widehat{\Sigma} - \Sigma ||_{\mathrm{op}}  > t   \right) \leq 2 \times  9^d \exp \left\{ -\frac{n}{2} \min \left\{ \frac{t^2}{(32\tau^2)^2}, \frac{t}{32\tau^2} \right\} \right\}. 
\end{align*}

Let $\sigma_d>\ldots>\sigma_1$ and $\widehat{\sigma}_d>\ldots>{\sigma}_1$ be respectively the ordered eigeinvalues of $\Sigma$ and $\widehat{\Sigma}$. Using Weyl's Theoreom \citet{weyl},  it holds:

\begin{align*}
 \mathbb{P} \left( \underset{1\leq k\leq d}{\max} |\widehat{\sigma}_k - \sigma_k | \; >\; t \right)\leq 2 \times  9^d \exp \left\{ -\frac{n}{2} \min \left\{ \frac{t^2}{(32\tau^2)^2}, \frac{t}{32\tau^2} \right\} \right\}. 
\end{align*}
\end{lemma}
\begin{proof}
 Let $N_{\rho}$ be an $\rho$-net of the sphere $\mathbb{S}^{d-1}$. Applying Lemma \ref{epsiloncovering} on $\widehat{\Sigma} - \Sigma$, for any $t,\rho>0$,  we have 

\begin{align*}
\mathbb{P} \left( ||\widehat{\Sigma} - \Sigma ||_{\text{op}}  > t \right) &\leq  \mathbb{P} \left(  \frac{1}{1-2\rho}  \; \underset{v \in N_{\rho}}{\max }  |v^\top (\widehat{\Sigma} - \Sigma) v| > t \right) \\& \leq  |N_{\rho}| \;  \mathbb{P} \left(  |v^\top (\widehat{\Sigma} - \Sigma) v|  > (1-2\rho)~t\right),
\end{align*}

where $ |N_{\rho}|$ stands for the cardinal of the set  $N_{\rho}$.  Noticing that $\widehat{\Sigma} = \frac{1}{n} \sum_{i=1}^{n} X_iX_i^\top$ is a sum of independent matrices we have 

\begin{align*}
v^\top (\widehat{\Sigma} - \Sigma) v = \frac{1}{n} \sum_{i=1}^{n} Z_i - \mathbb{E}Z_i,
\end{align*}
where $Z_i=(v^\top X_i)^2$ for every $1\leq i\leq n$ and $Z_i- \mathbb{E}Z_i$ are i.i.d  random variables that are  $((16\tau) ^2, 16\tau ^2)$ sub-exponential.

Choosing $\rho=1/4$, noticing that $N_{1/4}\leq 9^d$  and applying the sub-exponential tail bound lead to the desired result.


\end{proof}

\subsection{Affine-Invariance is not fulfilled by IRW in general}\label{exampleAI}

Here we provide an example of discrete distribution where IRW does not satisfy affine-invariance property.

Consider the discrete probability measure $P$ assigning the weight $1/3$ to the bivariate points in $\mathcal{D}_3=\{(-1,2), (3,3), (2,1) \}$ and let us compute the IRW depth of $x=(0,1)$ and $y=(3,2)$ relative to $P$. It is easy to see that the mappings  $u\in\mathbb{S}^1\mapsto D_{\mathrm{H,1}}(\langle u,\;  x\rangle , P_{u})$ and $u\in\mathbb{S}^1\mapsto D_{\mathrm{H,1}}(\langle u,\;  y\rangle , P_{u})$  take only two values, $0$ or $1/3$. Identifying $\mathbb{S}^1$ as $[0,2\pi[$, the univariate halfspace depth of $x$ relative to $P$ is then null for any $u \in[\pi/4, \pi/2]\cup [5\pi/4, 3\pi/2]$ and equal to $1/3$ if  $u$ belongs to the complementary set.  In addition, $D_{\mathrm{H,1}}(\langle u,\;  y\rangle , P_{u})$ is equal to $0$ for any $u\in [3\pi/4, \pi]\cup [7\pi/4, 2\pi]$ and equal to $1/3$ on the complementary set. One may easily check that $D_{\text{IRW}}(x,P)= D_{\text{IRW}}(y,P)= 0.25$ and the same rank would be then assigned to each point by the IRW depth.  Now, multiplying all ordinate values by $2$, which is an affine transformation, the univariate halfspace depth of $\widetilde{x}=(0,2)$ is now null for all $u$ in  $[\pi/8, \pi/2]\cup [9\pi/8, 3\pi/2]$ while it remains equal to $1/3$ on the complementary set of this region. The depth of $\widetilde{x}$ is thus  lower than $0.25$. On the other hand, the univariate depth of $\widetilde{y}=(3,4)$ is now null on  $[7\pi/8, \pi]\cup [15\pi/8, 2\pi]$ while it remains equal to $1/3$ on the complementary set of this interval. It follows that  $D_{\text{IRW}}(\widetilde{x})=5/24<0.25< 7/24= D_{\text{IRW}}(\widetilde{y})$.

\section{Technical Proofs of the Main Results}
We now prove the main results stated in the paper.
\subsection{Proof of Proposition \ref{proposition}}\label{aiirw:proof:prop}
\subsubsection{Affine-Invariance}
 Let $A \in \mathbb{R}^{d \times d}$ be a non singular matrix and $b\in \mathbb{R}^d$. Let $\Sigma_{\scriptscriptstyle X}$ and $\Sigma_{\scriptscriptstyle AX}$ the covariance matrix of $X$ and $AX$ respectively. Defines the Cholesky decomposition as $\Sigma_{\scriptscriptstyle X}=\Lambda_{\scriptscriptstyle X}\Lambda_{\scriptscriptstyle X}^\top$ and $\Sigma_{\scriptscriptstyle AX}= A \Lambda_{ \scriptscriptstyle X}\Lambda_{\scriptscriptstyle X}^\top A^\top= \Lambda_{\scriptscriptstyle AX}\Lambda_{\scriptscriptstyle  AX}^\top$. It holds:
 
\begin{align*}
D_{\scriptscriptstyle \text{AI-IRW}}(Ax+b, AX+b )&=\frac{1}{V_d} \int_{\mathbb{S}^{d-1}} D_{\mathrm{H,1}}( \langle \frac{\Lambda_{\scriptscriptstyle  AX+b}^{\scriptscriptstyle -\top }u }{||\Lambda_{\scriptscriptstyle  AX+b}^{\scriptscriptstyle -\top}u ||},Ax+b \rangle,\; \langle \frac{\Lambda_{\scriptscriptstyle  AX+b}^{\scriptscriptstyle -\top}u }{||\Lambda_{\scriptscriptstyle  AX+b}^{\scriptscriptstyle -\top}u ||},AX+b \rangle ) \; du \\&= \frac{1}{V_d} \int_{\mathbb{S}^{d-1}} D_{\mathrm{H,1}}( \langle \Lambda_{\scriptscriptstyle  AX+b}^{\scriptscriptstyle -\top }u ,Ax+b \rangle,\; \langle \Lambda_{\scriptscriptstyle  AX+b}^{\scriptscriptstyle -\top}u ,AX+b \rangle ) \; du \\&=\frac{1}{V_d} \int_{\mathbb{S}^{d-1}} D_{\mathrm{H,1}}( \langle \Lambda_{\scriptscriptstyle  AX}^{\scriptscriptstyle -\top }u ,Ax \rangle,\; \langle \Lambda_{\scriptscriptstyle  AX}^{\scriptscriptstyle -\top}u ,AX \rangle ) \; du  \\& = \frac{1}{V_d} \int_{\mathbb{S}^{d-1}} D_{\mathrm{H,1}}(\langle u, \Lambda_{\scriptscriptstyle  X}^{\scriptscriptstyle -1}x \rangle, \langle u, \Lambda_{\scriptscriptstyle  X}^{\scriptscriptstyle -1}X \rangle )\; du \\& = \frac{1}{V_d} \int_{\mathbb{S}^{d-1}} D_{\mathrm{H,1}}(\langle \frac{\Lambda_{\scriptscriptstyle  X}^{\scriptscriptstyle -\top} u}{||\Lambda_{\scriptscriptstyle  X}^{\scriptscriptstyle -\top} u||}, x \rangle, \langle \frac{\Lambda_{\scriptscriptstyle  X}^{\scriptscriptstyle -\top} u}{||\Lambda_{\scriptscriptstyle  X}^{\scriptscriptstyle -\top} u||}, X\rangle )\; du 
 \\& = D_{\scriptscriptstyle \text{AI-IRW}}(x, P).
\end{align*}
The same reasoning applies if the square matrix is given by the SVD decomposition.
\subsubsection{Proof of Maximality at the Center}
Assume that $P$ is halfspace symmetric about a unique $\beta$, i.e., $\mathbb{P} \left(X\in \mathcal{H}_{\beta}\right)\geq \frac{1}{2}$ for every closed halfspace $\mathcal{H}_{\beta}$ such that $\beta\in  \partial \mathcal{H}$ with $\partial \mathcal{H}$ the boundary of $\mathcal{H}$. Thus, it is easy to see that $D_{\scriptscriptstyle \text{AI-IRW}}(\beta,P)\geq \frac{1}{2}$.  The uniqueness of $\beta$ and the fact that  $D_{\scriptscriptstyle \text{AI-IRW}}$  is lower than $1/2$  for any element in $\mathbb{R}^d$ by definition imply that $$\beta= \underset{x \in \mathbb{R}^d }{\text{argsup }}D_{\scriptscriptstyle \text{AI-IRW}}(x,P).$$

\subsubsection{Vanishing at Infinity}

The proof is a particular case of the proof of theorem 1 in \citet{cuevas2009}. We detail it for the sake of clarity. Let $U$ be a random variable following $\omega_{d-1}$, the uniform measure on the unit sphere $\mathbb{S}^{d-1}$. Defines $V=\Sigma^{{\scriptscriptstyle - \top/2}} U/\vert\vert \Sigma^{{\scriptscriptstyle - \top/2}} U \vert\vert$ and $\nu_{d-1}$ its probability distribution. Let $\theta>0$ and $x\in \mathbb{R}^d$, then  $r(\theta):=\nu_{d-1} \{v: \;  \frac{|\langle v,x\rangle|}{||x||}\leq \theta \}$ goes to zero when $\theta \rightarrow 0$. For any $x\in \mathbb{R}^d _{ \backslash \{0\}}$, we have

\begin{align*}
 D_{\scriptscriptstyle \text{AI-IRW}}(x, P)&= \int_{\mathbb{R}^{d}} \min \left\{ F_v(v^\top x), 1-F_v(v^\top  x) \right\}  \; d\nu_{d-1} (v) \\& \leq \int_{\mathbb{R}^{d}} \mathbb{I} \left\{v: \; \frac{|\langle v,x \rangle|}{||x||} \leq \theta \right\} \; d\nu_{d-1} (v) \\& + \int_{\mathbb{R}^{d}} F_v(v^\top x) \; \mathbb{I} \left\{v: \; \frac{|\langle v,  x \rangle|}{||x||} > \theta, \; \langle v, x \rangle\leq 0 \right\} \; d\nu_{d-1} (v) \\& \; \; + \int_{\mathbb{R}^{d}} (1-F_v(v^\top  x) ) \; \mathbb{I} \left\{v: \; \frac{|\langle v,x \rangle|}{||x||} > \theta, \; \langle v,x \rangle > 0 \right\} \; d\nu_{d-1} (v) \\& \leq r(\theta) +\int_{\mathbb{R}^{d}} F_v(-\theta ||x||) \; \mathbb{I} \left\{v: \; \frac{|\langle v,x \rangle|}{||x||} > \theta, \; \langle v,x \rangle\leq 0 \right\} \; d\nu_{d-1} (v) \\& \; \;  +\int_{\mathbb{R}^{d}} (1-F_v(\theta ||x||) ) \; \mathbb{I} \left\{v: \; \frac{|\langle v, x \rangle|}{||x||} > \theta, \; \langle v,x \rangle > 0 \right\} \; d\nu_{d-1} (v).
\end{align*}

Now, when $||x||\rightarrow \infty$, the dominated convergence theorem ensures that $$\lim \underset{||x|| \rightarrow \infty}{\sup} D_{\scriptscriptstyle \text{AI-IRW}}(x, P) \leq r(\theta) \underset{\theta \rightarrow 0}{\rightarrow} 0.$$

\subsubsection{Decreasing along Rays}

The proof is a slight modification of the proof of Assertion (iii) of Theorem 2 in \citet{IRW}. Details are left to the reader.

\subsubsection{Continuity}
For any $P\in \mathcal{P}(\mathbb{R}^d)$, the continuity of the inner product and the cdf ensure continuity of  $D_{\mathrm{H}}(v^\top x, v^\top X)$ for any $v \in \mathbb{S}^{d-1}$. Therefore, the continuity of $x\mapsto D_{\scriptscriptstyle \text{AI-IRW}}(x, P)$ follows from dominated convergence.



\subsection{Proof of Theorem \ref{theorem}}\label{aiirw:proof:theorem}
We now prove the main results of the paper.
\subsubsection{Assertion (i)}
Introducing terms and using triangle inequality, it holds:

\begin{align*}
\underset{x \in \mathbb{R}^d}{\sup } \;  \Big \lvert \widehat{D}_{\scriptscriptstyle \text{AI-IRW}}(x) -    D_{\scriptscriptstyle \text{AI-IRW}}(x, P) \Big \rvert&\leq \underbrace{\underset{x \in \mathbb{R}^d}{\sup } \;  \Big \lvert \widehat{F}_{\widehat{V}}(\widehat{V}^\top x) -F_{\widehat{V}}(\widehat{V}^\top x) \Big \rvert}_{(1)}   + \underbrace{\underset{x \in \mathbb{R}^d}{\sup } \;\Big \lvert F_{\widehat{V}}(\widehat{V}^\top x) -F_{V}(V^\top x) \Big \rvert }_{(2)}. 
\end{align*}

 Now, the first term (1) can be controlled using the  bound for the deviations of  halfspace depth deferred in  Lemma \ref{tukey}. Thus, for any $t>0$ it holds: 

\begin{align}\label{halfspacebound}
\mathbb{P} \left( \underset{x \in \mathbb{R}^d}{\sup } \;  \Big \lvert \widehat{F}_{\widehat{V}}(\widehat{V}^\top x) -F_{\widehat{V}}(\widehat{V}^\top x) \Big \rvert  \; >\; t/2 \right) &\leq \mathbb{P} \left( \underset{\substack{y\in \mathbb{R}^d \\ u \in \mathbb{S}^{d-1}}}{\sup } \; \Big \lvert \widehat{F}_{u}(u ^\top y) -F_{u}(u^\top y) \Big \rvert \; > \;t/2 \right) \nonumber \\& \leq \frac{6 (2n)^{d+1}}{(d+1)!}\exp(-nt^2/32).
\end{align}

 The second term (2) relies on the influence of the deviations of the sample covariance matrix. First remark that:

\begin{align*}
\underset{x\in \mathbb{R}^d}{\sup } \;\Big \lvert  F_{\widehat{V}}(\widehat{V}^\top x) - F_{V}(V^\top x)  \Big \rvert & \leq \underset{\substack{x\in \mathbb{R}^d \\ u \in \mathbb{S}^{d-1}}}{\sup } \;  \Big  \lvert  \mathbb{P} \left( \langle  \frac{\widehat{\Sigma}^{\scriptscriptstyle -\top/2} u}{||\widehat{\Sigma}^{\scriptscriptstyle -\top/2} u||},  X-x \rangle  \leq 0 \;  \Big\vert \; \mathcal{S}_n\right) \\& \quad  - \mathbb{P} \left( \langle \frac{\Sigma^{\scriptscriptstyle -\top/2} u}{||\Sigma^{\scriptscriptstyle -\top/2} u|| },  X-x \rangle   \leq 0\right) \Big \rvert. 
\end{align*}


Now, since $X$ is radially Lipschitz continuous, we have: 
{\small
\begin{align*}
\Bigg \lvert  \mathbb{P} \left( \langle  \frac{\widehat{\Sigma}^{\scriptscriptstyle -\top/2} u}{||\widehat{\Sigma}^{\scriptscriptstyle -\top/2} u||},  X-x \rangle  \leq 0 \;  \Big \vert \; \mathcal{S}_n\right) &- \mathbb{P} \left( \langle \frac{\Sigma^{\scriptscriptstyle -\top/2} u}{||\Sigma^{\scriptscriptstyle -\top/2} u|| },  X-x \rangle   \leq 0\right)  \Bigg \rvert   \leq L_R \left| \left|\frac{\widehat{\Sigma}^{\scriptscriptstyle -\top/2} u}{||\widehat{\Sigma}^{\scriptscriptstyle -\top/2} u ||} - \frac{\Sigma^{\scriptscriptstyle -\top/2} u}{ ||\Sigma^{\scriptscriptstyle -\top/2} u ||} \right| \right|. 
\end{align*}
}%


 Introducing terms and using triangle inequality leads to:

\begin{align*}
\left| \left|\frac{\widehat{\Sigma}^{\scriptscriptstyle -\top/2} u}{||\widehat{\Sigma}^{\scriptscriptstyle -\top/2} u ||} - \frac{\Sigma^{\scriptscriptstyle -\top/2} u}{ ||\Sigma^{\scriptscriptstyle -\top/2}  u ||} \right| \right|  &\leq \frac{||\widehat{\Sigma}^{\scriptscriptstyle -1/2} - \Sigma^{\scriptscriptstyle -1/2} ||_{\text{op}}}{||\Sigma^{\scriptscriptstyle -1/2} u ||} + || \widehat{\Sigma}^{\scriptscriptstyle -1/2}  u|| \left( \frac{1}{||\widehat{\Sigma}^{\scriptscriptstyle -1/2} u ||} - \frac{1}{||\Sigma^{\scriptscriptstyle -1/2} u ||}\right) \\& \leq \frac{2 ||\widehat{\Sigma}^{\scriptscriptstyle -1/2} - \Sigma^{\scriptscriptstyle -1/2} ||_{\text{op}}}{||\Sigma^{\scriptscriptstyle -1/2} u ||},
\end{align*}



yielding:

\begin{align}\label{bornetotal}
\underset{x\in \mathbb{R}^d}{\sup } \; \Big \lvert  F_{\widehat{V}}(\widehat{V}^\top x) - F_{V}(V^\top x)  \Big \rvert \leq \frac{2 L_R}{||\Sigma^{\scriptscriptstyle -1/2}  ||_{\text{op}}} \; ||\widehat{\Sigma}^{\scriptscriptstyle -1/2} - \Sigma^{\scriptscriptstyle -1/2} ||_{\text{op}}. 
\end{align}

Assume that $ODO^\top$ and $\widehat{O}\widehat{D}\widehat{O}^\top$ are  the eigenvalues decomposition  of $\Sigma$ and  $\widehat{\Sigma}$ in orthonormal bases. Thus, thanks to Lemma \ref{lemma:covariance:decompo}, we have:

%
%

\begin{align*}
 ||\widehat{\Sigma}^{\scriptscriptstyle -1/2} - \Sigma^{\scriptscriptstyle -1/2} ||_{\text{op}} &\leq ||\Sigma^{\scriptscriptstyle -1/2}  ||_{\text{op}} \left( ||\widehat{D}^{1/2} -D^{1/2} ||_{\text{op}} \; ||\widehat{D}^{-1/2} ||_{\text{op}} +  || \widehat{O} - O ||_{\text{op}} \right)\nonumber \\& 
\end{align*}
Now, since 
$\underset{k\leq d}{\min} \; \sqrt{\widehat{\sigma}_k} \geq  \sqrt{\varepsilon} - \underset{k\leq d}{\max}\; |\sqrt{\widehat{\sigma}_k} - \sqrt{\sigma}_k| $ and  $\underset{k\leq d}{\max}\; |\sqrt{\widehat{\sigma}_k} - \sqrt{\sigma}_k| \leq \frac{1}{\sqrt{\varepsilon}}\underset{1 \leq k\leq d}{\max } \; |\widehat{\sigma}_k - \sigma_k |$, using Weyl's inequality leads to:

\begin{align*}\label{event}
\underset{x\in \mathbb{R}^d}{\sup } \; \Big \lvert  F_{\widehat{V}}(\widehat{V}^\top x) - F_{V}(V^\top x)  \Big \rvert&\leq  2L_R  \left( \frac{||\widehat{\Sigma} - \Sigma ||_{\text{op}}}{\; \varepsilon- ||\widehat{\Sigma} - \Sigma ||_{\text{op}}} + || \widehat{O} - O ||_{\text{op}} \right).
\end{align*}

By $\mathcal{A}_{\xi}$  it is denoted the event  $\mathcal{A}_{\xi} =\left\{ ||\widehat{\Sigma} - \Sigma ||_{\text{op}} < \varepsilon - \xi \right\}$ for any $\xi \in [0,\varepsilon )$. Using union bound and  combining (\ref{bornetotal}) with the previous equation,  for any $t>0$  and $\xi \in (0,\varepsilon )$ it holds:
\begin{align*}
\mathbb{ P} \left( \underset{x\in \mathbb{R}^d}{\sup } \;\Big \lvert  F_{\widehat{V}}(\widehat{V}^\top x) - F_{V}(V^\top x)  \Big \rvert \; > \; t/2 \right) &\leq \mathbb{P} \left( \frac{2L_R}{\xi}||\widehat{\Sigma} - \Sigma ||_{\text{op}} > \; t/4\right)+ \mathbb{P} \left( \mathcal{A}^c_{\xi}\right) \\& \quad + \mathbb{P} \left( 2L_R~||\widehat{O}- O ||_{\text{op}} > \;  t/4 \right),
\end{align*}
where $\mathcal{A}^c_{\xi}$ stands for the complementary event of $\mathcal{A}_{\xi}$. Applying Lemma  \ref{lemma:covariance:bounds} gives:

\begin{equation}\label{eq1}
\mathbb{P} \left(||\widehat{\Sigma} - \Sigma ||_{\text{op}} > \frac{\xi t}{8L_R}\right) \leq 2 \times  9^d \exp \left\{ -\frac{n}{2} \min \left\{ \frac{(\xi  t)^2}{(256 L_R\tau^2)^2}, \frac{\xi  t}{256 L_R\tau^2} \right\} \right\},
\end{equation}

and 
\begin{equation}\label{eq2}
\mathbb{P} \left( \mathcal{A}^c_{\xi}\right) \leq 2 \times  9^d \exp \left\{ -\frac{n}{2} \min \left\{ \frac{\left(\varepsilon - \xi\right)^2}{(32\tau^2)^2}, \frac{\varepsilon - \xi}{32\tau^2} \right\} \right\}.
\end{equation}

\vspace*{0.2cm}

Furthermore, it is easy to see that $||\widehat{O}-O||_{\text{op}} \leq \sqrt{d}~ \underset{ k \leq d}{\max}\;  || \widehat{O}_k - O_k||$  
where $O_k$ is the $k$-th column of the matrix $O$. Let $\gamma$ be the minimum eigengap, following a variant of the Davis-Kahan theorem \citep{daviskahan} (see corollary 1 in \citet{wangdaviskahan}),  it holds:
\begin{align*}\label{eq:vectors}
||\widehat{O}-O||_{\text{op}} & \leq \frac{2\sqrt{2d}  || \widehat{\Sigma}-\Sigma ||_{\text{op}}}{\gamma}  .
\end{align*}

Using  Lemma  \ref{lemma:covariance:bounds} again leads to:

\begin{equation}\label{eq3}
\mathbb{P} \left(  \frac{4L_R\sqrt{2d}  || \widehat{\Sigma}-\Sigma ||_{\text{op}}}{\gamma} \; > \; t/4 \right) \leq 2 \times  9^d \exp \left\{ -\frac{n}{2} \min \left\{ \frac{(  \gamma t)^2}{(512 L_R\sqrt{2d} \tau^2)^2}, \frac{\gamma t}{512 L_R \sqrt{2d} \tau^2} \right\} \right\}.
\end{equation}


Combining  (\ref{eq1}), (\ref{eq2}) and (\ref{eq3}) it holds:

\begin{equation*}
\mathbb{ P} \left( \underset{x\in \mathbb{R}^d}{\sup } \;\Big \lvert  F_{\widehat{V}}(\widehat{V}^\top x) - F_{V}(V^\top x)  \Big \rvert \; > \; t/2 \right) \leq 6 \times 9^d \exp \left(-\frac{n}{2} \min \left\{ \left(\kappa t\right) ^2, \kappa t \right\} \right),
\end{equation*}
 for any $t\leq (\varepsilon- \xi)/ (32\tau^2\kappa)$  where $\kappa= \dfrac{1}{256 L_R \tau^2}\left( \xi \wedge \dfrac{\gamma  }{2\sqrt{2d} } \right)$. Finally,  for any $t\leq (\varepsilon- \xi)/ (32\tau^2\kappa)$ it holds: 

{\small
\begin{equation}\label{final}
\mathbb{P} \left( \underset{x \in \mathbb{R}^d}{\sup } \;  \Big \lvert \widehat{D}_{\scriptscriptstyle \text{AI-IRW}}(x) -    D_{\scriptscriptstyle \text{AI-IRW}}(x, P) \Big \rvert \;>\; t\right)\leq 6.9^d\exp \left(-\frac{n}{2} \min \left\{ \left(\kappa t\right) ^2, \kappa t \right\} \right) + 
\frac{6 (2n)^{d+1}}{(d+1)!}\exp(-nt^2/32).
\end{equation}
}%
Bounding each term in the right side by $\delta/2$ and reverting the equation lead to the desired result.

\subsubsection{Assertion (ii)} \label{assertion2}
Let $\mathcal{B}_r$ a centered  ball of $\mathbb{R}^d$ with radius $r>0$ and  assume that $X$ satisfies assumption 2 for any $x\in \mathcal{B}_r$. Introducing terms and using triangle inequality, it holds:

\begin{align*}
 \underset{x \in \mathcal{B}_r}{\sup } \;  \Big \lvert \widetilde{D}_{\scriptscriptstyle \text{AI-IRW}}^{\text{MC}}(x) -    D_{\scriptscriptstyle \text{AI-IRW}}(x, P) \Big \rvert & \leq \underbrace{\underset{x \in \mathbb{R}^d}{\sup } \;  \Big \lvert \widehat{D}_{\scriptscriptstyle \text{AI-IRW}}(x) -    D_{\scriptscriptstyle \text{AI-IRW}}(x, P) \Big \rvert}_{(1)}  \\& + \underbrace{ \underset{x \in \mathcal{B}_r}{\sup }  \Big \lvert D_{\scriptscriptstyle \text{AI-IRW}}^{\text{MC}}(x, P) -    D_{\scriptscriptstyle \text{AI-IRW}}(x, P) \Big \rvert}_{(2)}
\end{align*}

The first term (1) can be bounded using assertion (i) while controlling the approximation term (2) relies on classical chaining arguments.  As the function $z\mapsto \min(z,1-z)$ is 1-Lipschitz for any $z\in (0,1)$ and by triangle inequality,  for any $y$ in $\mathcal{B}_r$ we have:
\begin{align*}
 \Big \lvert D_{\scriptscriptstyle \text{AI-IRW}}^{\text{MC}}(y, P) -    D_{\scriptscriptstyle \text{AI-IRW}}(y, P) \Big \rvert\leq \frac{1}{m}\sum_{j=1}^{m}  \Big \lvert \mathbb{P}\left\{ \langle  V_j,   y \rangle \; \big\vert \;  V_j\right\} - \mathbb{P}\left\{ \langle V,  y \rangle \right\} \Big \rvert.
\end{align*}

Since it is an average of bounded and  i.i.d random variables, combining Hoeffding inequality and union bound, for any $t>0$ and any $y$ in $\mathcal{B}_r$ it holds:
\begin{align}\label{hoeff}
\mathbb{P} \left( \Big \lvert D_{\scriptscriptstyle \text{AI-IRW}}^{\text{MC}}(y, P) -    D_{\scriptscriptstyle \text{AI-IRW}}(y, P) \Big \rvert \; >\; t/2 \right)\leq 2\exp\left( -mt^2/2\right).
\end{align}

As $X$ is uniformly continuous Lipschitz in projection for any $u\in \mathbb{S}^{d-1}$, observe that $\forall (x,y) \in \mathcal{B}_r^2$  it holds:

\begin{align}\label{decompo}
 \Big \lvert D_{\scriptscriptstyle \text{AI-IRW}}^{\text{MC}}(x, P) -    D_{\scriptscriptstyle \text{AI-IRW}}(x, P) \Big \rvert & \leq  \Big \lvert D_{\scriptscriptstyle \text{AI-IRW}}^{\text{MC}}(x, P) -    D_{\scriptscriptstyle \text{AI-IRW}}^{\text{MC}}(y, P) \Big \rvert +  \Big \lvert D_{\scriptscriptstyle \text{AI-IRW}}^{\text{MC}}(y, P) -    D_{\scriptscriptstyle \text{AI-IRW}}(y, P) \Big \rvert \nonumber \\& \; +  \Big \lvert D_{\scriptscriptstyle \text{AI-IRW}}(x, P) -    D_{\scriptscriptstyle \text{AI-IRW}}(y, P) \Big \rvert \nonumber \\& \leq   2 L_p~  ||x-y||+ \Big \lvert D_{\scriptscriptstyle \text{AI-IRW}}^{\text{MC}}(y, P) -    D_{\scriptscriptstyle \text{AI-IRW}}(y, P) \Big \rvert.
\end{align}

Now let $\zeta>0$ and $y_1,\ldots,y_{\mathcal{N}(\zeta, \mathcal{B}_r, ||.||_2)}$ be a $\zeta$-coverage of $\mathcal{B}_r$ with respect to $||.||_2$. We have: 

\begin{align}\label{covering}
\log \left( \mathcal{N}(\zeta,\mathcal{B}_r, ||.||_2)\right)  \leq d \log \left(3r/\zeta \right). 
\end{align}

Set $\mathcal{N}=\mathcal{N}\left(\zeta, \mathcal{B}_r, ||.||_2)\right)$ for simplicity. There exists $\ell \leq \mathcal{N}$ such that $||x-y_{\ell}||_2 \leq \zeta$. Thus, Eq. (\ref{decompo}) leads to

\begin{align*}
\Big \lvert D_{\scriptscriptstyle \text{AI-IRW}}^{\text{MC}}(x, P) -    D_{\scriptscriptstyle \text{AI-IRW}}(x, P) \Big \rvert  \leq 2 L_p  \; \zeta +\Big \lvert D_{\scriptscriptstyle \text{AI-IRW}}^{\text{MC}}(y_{\ell}, P) -    D_{\scriptscriptstyle \text{AI-IRW}}(y_{\ell}, P) \Big \rvert.
\end{align*}

Applying  Eq. (\ref{hoeff}) to every $y_{\ell}$ and the union bound, for any $t>0$, we get:

\begin{align*}
\mathbb{P} \left( \underset{\ell \leq \mathcal{N}}{\sup }\;   \Big \lvert D_{\scriptscriptstyle \text{AI-IRW}}^{\text{MC}}(y_{\ell}, P) -    D_{\scriptscriptstyle \text{AI-IRW}}(y_{\ell}, P) \Big \rvert \; >\;  t/2\right) \leq 2\mathcal{N}\exp\left( -mt^2/2\right),
\end{align*}

yielding:
\begin{align*}
\mathbb{P} \left( \underset{x \in \mathcal{B}_r}{\sup }\;   \Big \lvert D_{\scriptscriptstyle \text{AI-IRW}}^{\text{MC}}(x, P) -    D_{\scriptscriptstyle \text{AI-IRW}}(x, P) \Big \rvert \; >\; t/2\right)\leq 2\mathcal{N}\exp\left( -2m \left( t/2 - 2L_p \zeta\right)^2\right).
\end{align*}

Using Eq. \eqref{final}, the union bound and \eqref{covering}, we obtain:

\begin{align*}
\mathbb{P}\left( \underset{x \in \mathcal{B}_r}{\sup } \;  \Big \lvert \widetilde{D}_{\scriptscriptstyle \text{AI-IRW}}^{\text{MC}}(x) -    D_{\scriptscriptstyle \text{AI-IRW}}(x, P) \Big \rvert \; >\; t \right) &\leq \mathbb{P}\left( \underset{x \in \mathcal{B}_r}{\sup } \;  \Big \lvert \widehat{D}_{\scriptscriptstyle \text{AI-IRW}}(x) -    D_{\scriptscriptstyle \text{AI-IRW}}(x, P) \Big \rvert \; >\; t/2\right) \\& \quad  + \mathbb{P} \left( \underset{x \in \mathcal{B}_r}{\sup }\;   \Big \lvert D_{\scriptscriptstyle \text{AI-IRW}}^{\text{MC}}(x, P) -    D_{\scriptscriptstyle \text{AI-IRW}}(x, P) \Big \rvert \; >\; t/2\right) \\& \leq 6.9^d\exp \left(-\frac{n}{2} \min \left\{ \left(\kappa t/2\right) ^2, \kappa t/2 \right\} \right) \\& \quad + 
\frac{6 (2n)^{d+1}}{(d+1)!}\exp(-nt^2/128) \\& \quad + 2\left(\frac{3r}{\zeta}\right)^d\exp\left( -2m \left( t/2 - 2L_p \zeta\right)^2\right) .
\end{align*}

Choosing $\zeta \sim m^{-1}$, bounding each term on the right-hand side by $\delta/3$ and reverting the previous equation lead to the desired result.


\subsection{Geometrical Results on the Lipschitz Constants of Assumptions \ref{as:2} and \ref{as:3}}

\begin{lemma} \label{lemma:argball}
Let $r>0$ and denote by $V_{d,r}$ the volume  of the d-ball $\mathcal{B}(0,r)$. Assume that $X$ takes its values in $\mathcal{B}(0,r)$ and has  an $M$- bounded density w.r.t. the Lebesgue measure $\lambda$. The r.v. $X$ is uniformly \textsc{Radially Lipschitz Continuous} with constant $L_R=M V_{d,r}$.
\end{lemma}

\begin{figure}[!h]
\begin{center}
\includegraphics[scale=0.4]{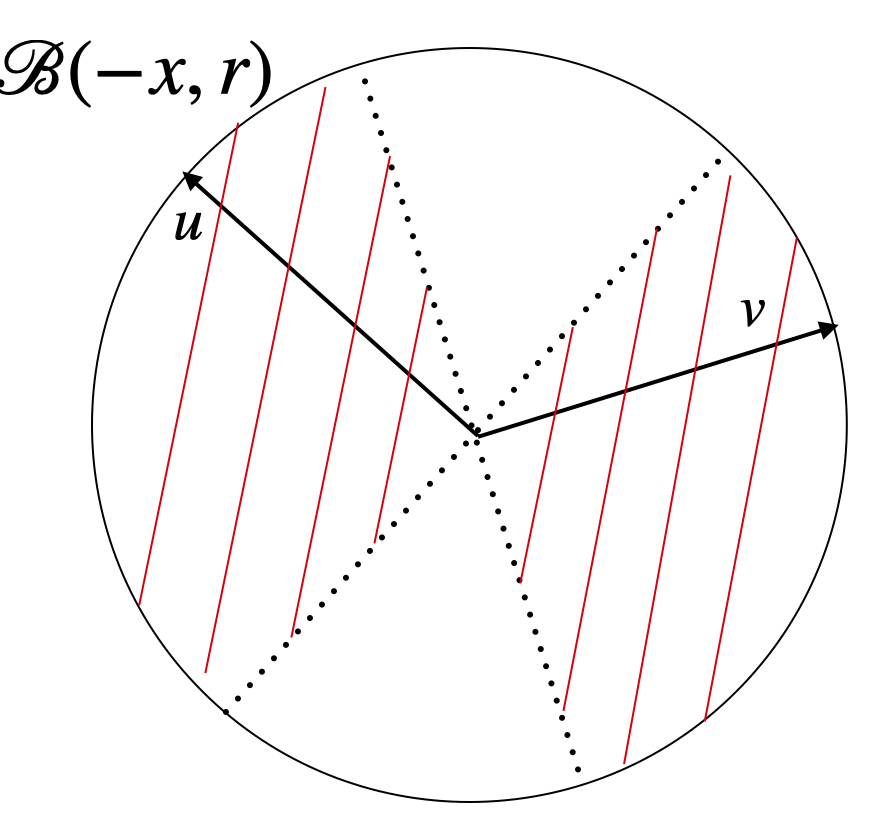}
\end{center}
\caption{Illustration of the set $\mathcal{E}_{u,x,y} $ in $\mathbb{R}^2$. It corresponds to the portion of $\mathcal{B}(-x,r)$ hatched in red.}
\label{figure:argball}
\end{figure}

\begin{proof}
Let $x \in \mathbb{R}^d$. By $||.||_g$ it means the geodesic norm on the unit sphere of $\mathbb{R}^d$. It holds:

\begin{align*}
|\phi(u,x)- \phi(v,x)| &\leq \mathbb{P} \left\{  X \in \mathcal{B}(0,r):  \quad   \langle  u  ,  X-x \rangle \text{ and }  \langle v ,  X-x \rangle \text{ are of opposite sign}\right\}\\&
 \leq  M \;  \lambda \left\{z \in \mathcal{B}(-x,r):  \quad   \langle u ,  z \rangle \text{ and }  \langle v  ,  z \rangle \text{ are of opposite sign} \right\} \\& \overset{(i)}{\leq}   \; M \; V_{d,r} \times \frac{2}{\pi} \; \text{arccos} \left(  \langle  u,  v \rangle \right) \\& =M \;  V_{d,r} \times \frac{2}{\pi} \; ||u-v||_g \\& \leq M \;  V_{d,r}\; ||u-v||, 
\end{align*}

Where (i) arises from the fact that the volume of $\mathcal{E}_{u,x,y}=\{z \in \mathcal{B}(-x,r):  \quad   \langle u  ,  z \rangle \text{ and }  \langle v  ,  z \rangle \text{ are of opposite sign}  \}$ is the volume of two cones of angle $||u-v||_g$, as depicted in Figure~\ref{figure:argball}.
\end{proof}

\begin{figure}[!h]
\begin{center}
\includegraphics[scale=0.4]{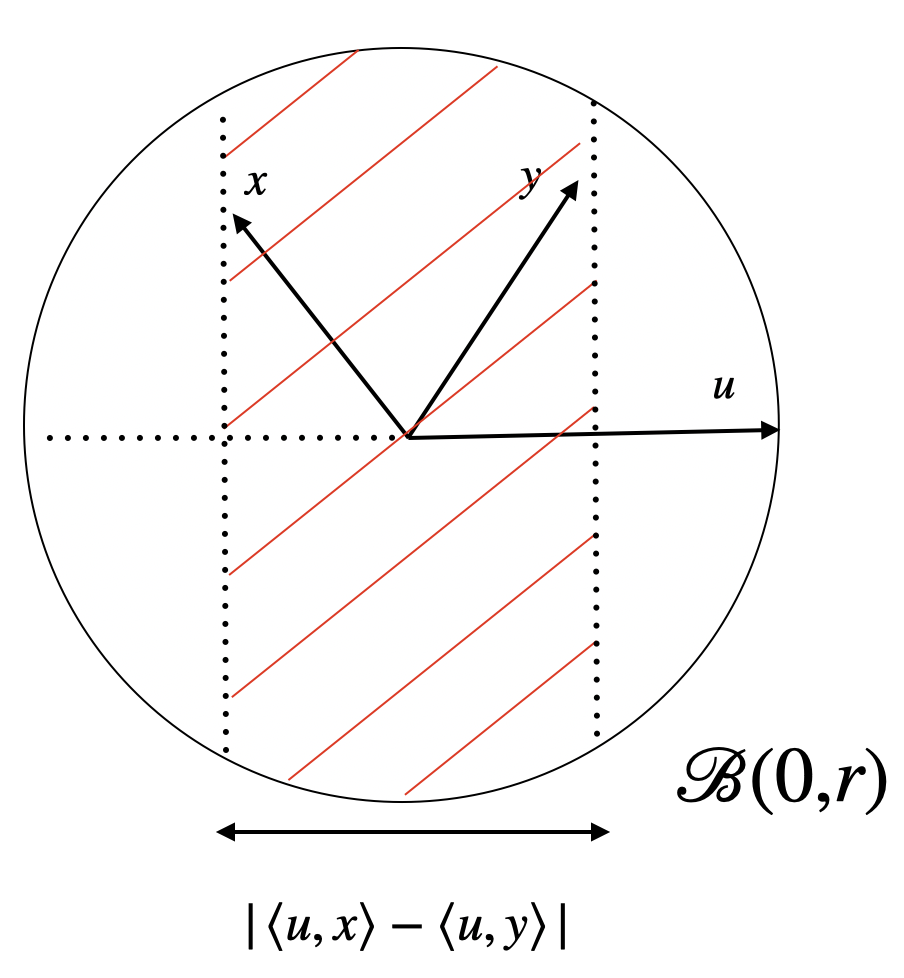}
\end{center}
\caption{Illustration of the set $\mathcal{F}_{u,x,y}$ in $\mathbb{R}^2$. It corresponds to the portion  of $\mathcal{B}(0,r)$ hatched in red. }
\label{figure:argball2}
\end{figure}

\begin{lemma} \label{lemma:argball2}
Let $r>0$ and  assume that $X$ takes its values in $\mathcal{B}(0,r)$ and has $M$-bounded density w.r.t. the Lebesgue measure $\lambda$. Thus $X$ is uniformly \textsc{Lipschitz continuous in projection} with constant $L_p=M V_{d-1,r}$.
\end{lemma}
\begin{proof}
Let $u\in \mathbb{S}^{d-1}$.  By $||.||_g$ it means the geodesic norm on the unit sphere of $\mathbb{R}^d$. It holds:

\begin{align*}
|\phi(u,x)- \phi(u,y)| &\leq \mathbb{P} \left\{  X \in \mathcal{B}(0,r):  \quad   \langle  u  ,  X-x \rangle \text{ and }  \langle u ,  X-y \rangle \text{ are of opposite sign}\right\}\\&  \leq M \; \lambda \left\{z \in \mathcal{B}(0,r):  \quad   \langle   u ,  z-x \rangle \text{ and }  \langle  u  , z-y \rangle \text{ are of opposite sign} \right\} \\& \overset{(i)}{\leq} M\;  V_{d-1,r}  \times   | \langle u,x \rangle -\langle u,y\rangle  |\\& \leq M\;  V_{d-1,r} \;  ||x-y||.
\end{align*}

Where (i) arises from the fact that we encompass $\mathcal{F}_{u,x,y}$ by an hyper-cylinder of length $| \langle u,x \rangle -\langle u,y\rangle  |$ where $\mathcal{F}_{u,x,y} =\{z \in \mathcal{B}(0,r):  \quad   \langle   u , z-x \rangle \text{ and }  \langle  u , z-y \rangle \text{ are of opposite sign}  \}$, as illustrated in Figure~\ref{figure:argball2}.
\end{proof}

\subsection{Finite-Sample Analysis of the IRW depth} \label{irwbound}

A finite sample analysis on IRW can be derived from our results on AI-IRW as it is described in the next corollary.

\begin{corollary} Suppose that the distribution $P$ of the r.v. X satisfies Assumptions \ref{as:2} and \ref{as:3}. Then, for any $\delta \in (0,1)$, it holds:

\begin{align*}
 \underset{x \in \mathcal{B}_r}{\sup } \;  \Big \lvert \widehat{D}_{\scriptscriptstyle \text{IRW}}^{\text{MC}}(x) -    D_{\scriptscriptstyle \text{IRW}}(x, P) \Big \rvert & \leq \sqrt{\dfrac{8\log(\Theta/\delta)}{n}} + 2\sqrt{\frac{d \log \left(3rm\right) + \log(6 /\delta)}{8m}}
 +\frac{2 L_p }{m},
\end{align*}
where  $\Theta=12 (2n)^{d+1}/(d+1)!$.
\end{corollary}

\begin{proof}
First notice that:
$$
 \underset{x \in \mathcal{B}_r}{\sup } \;  \Big \lvert \widehat{D}_{\scriptscriptstyle \text{IRW}}^{\text{MC}}(x) -    D_{\scriptscriptstyle \text{IRW}}(x, P) \Big \rvert  \leq \underbrace{\underset{x \in \mathbb{R}^d}{\sup } \;  \Big \lvert \widehat{D}_{\scriptscriptstyle \text{IRW}}(x) -    D_{\scriptscriptstyle \text{IRW}}(x, P) \Big \rvert}_{(1)}   + \underbrace{ \underset{x \in \mathcal{B}_r}{\sup }  \Big \lvert D_{\scriptscriptstyle \text{IRW}}^{\text{MC}}(x, P) -    D_{\scriptscriptstyle \text{IRW}}(x, P) \Big \rvert}_{(2)}.
 $$
 Now, the first term (1) can be controlled using the bound for the deviations of  Halfspace Depth deferred in  Lemma \ref{tukey}. Thus, for any $t>0$, it holds: 

\begin{align}\label{halfspacebound2}
\mathbb{P}\left( \underset{x \in \mathbb{R}^d}{\sup } \;  \Big \lvert \widehat{D}_{\scriptscriptstyle \text{IRW}}(x) -    D_{\scriptscriptstyle \text{IRW}}(x, P) \Big \rvert \; > \; t/2 \right) \leq \frac{6 (2n)^{d+1}}{(d+1)!}\exp(-nt^2/32).
\end{align}

The second term can be bounded following the same reasoning than for the Monte-Carlo approximated term of AI-IRW described in Section ~\ref{assertion2}. Thus, with the same notations, for any $t>0$, we have: 

\begin{align}\label{approx}
\mathbb{P}\left(  \underset{x \in \mathcal{B}_r}{\sup }  \Big \lvert D_{\scriptscriptstyle \text{IRW}}^{\text{MC}}(x, P) -    D_{\scriptscriptstyle \text{IRW}}(x, P) \Big \rvert \; > \; t/2 \right)\leq 2\mathcal{N}\exp\left( -2m \left( t/2 - 2L_p \zeta\right)^2\right).
\end{align}

Using \Cref{halfspacebound2} and \Cref{approx}, one gets:

\begin{align*}
\mathbb{P}\left( \underset{x \in \mathcal{B}_r}{\sup } \;  \Big \lvert \widehat{D}_{\scriptscriptstyle \text{IRW}}^{\text{MC}}(x) -    D_{\scriptscriptstyle \text{IRW}}(x, P) \Big \rvert \; >\; t \right) &\leq \mathbb{P}\left( \underset{x \in \mathcal{B}_r}{\sup } \;  \Big \lvert \widehat{D}_{\scriptscriptstyle \text{IRW}}(x) -    D_{\scriptscriptstyle \text{IRW}}(x, P) \Big \rvert \; >\; t/2\right) \\& \quad  + \mathbb{P} \left( \underset{x \in \mathcal{B}_r}{\sup }\;   \Big \lvert D_{\scriptscriptstyle \text{IRW}}^{\text{MC}}(x, P) -    D_{\scriptscriptstyle \text{IRW}}(x, P) \Big \rvert \; >\; t/2\right) \\& \leq 
\frac{6 (2n)^{d+1}}{(d+1)!}\exp(-nt^2/32)  + 2\left(\frac{3r}{\zeta}\right)^d\exp\left( -2m \left( t/2 - 2L_p \zeta\right)^2\right) .
\end{align*}

Choosing $\zeta \sim m^{-1}$, bounding each term on the right-hand side by $\delta/2$ and reverting the previous equation lead to the desired result.

\end{proof}

\section{The Approximation Algorithm}\label{algoapprox}

In this section, we display the approximation algorithm of the AI-IRW depth, see Algorithm \ref{algo::AIIRW}.

\begin{algorithm}[H]
\caption{Approximation of the AI-IRW depth}
\textit{Initialization:} The number of projections $m$.
      \begin{algorithmic}[1]
      \STATE Construct $\mathbf{U}\in \mathbb{R}^{d \times m}$ by sampling uniformly $m$ vectors  $U_1,\ldots,U_m$ in $\mathbb{S}^{d-1}$
      \STATE Compute $\widehat{\Sigma}$ using any estimator
      \STATE Perform Cholesky or SVD on $\widehat{\Sigma}$  to obtain $\widehat{\Sigma}^{-1/2}$
      \STATE Compute $\mathbf{V}= \widehat{\Sigma}^{-1/2}\mathbf{U}/ ||\widehat{\Sigma}^{-1/2}\mathbf{U} ||$
      \STATE Compute $\mathbf{M}=\mathbf{XV}$
      \STATE Compute the rank value $\sigma(i,j)$, the rank of index $i$ in $\mathbf{M}_{:,j}$ for every $i\leq n$ and $j\leq m$
      \STATE Set $D_i=\frac{1}{m} \sum_{j=1}^{m} \; \; \sigma(i,j)$ for every $i\leq n$\\
      
 \textbf{Output}: $D$
      \end{algorithmic}
      \label{algo::AIIRW}
\end{algorithm}

\section{Additional Experiments}

\subsection{Illustration of (non) Affine-Invariance}\label{figAI}

\begin{figure}[!h]
\begin{center}
\begin{tabular}{cc}
\includegraphics[scale=0.4, trim= 2cm 0cm 0cm 0cm]{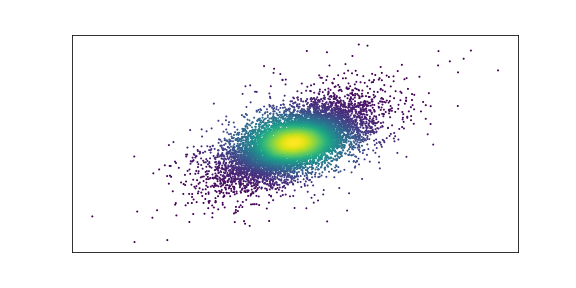}&
\includegraphics[scale=0.4, trim=2cm 0cm 0cm 0cm]{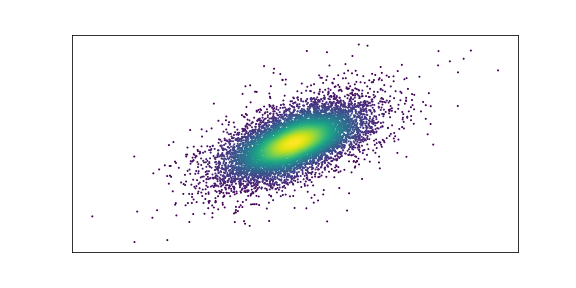}
\end{tabular}
\end{center}
\caption{The IRW depth (left) and the AI-IRW (right) depth on a Student-10 distribution. The darker the point, the lower the depth.}
\label{fig:gaussian}
\end{figure}

The Figure~\ref{fig:gaussian} illustrates the non affine-invariance of the IRW and the affine-invariance of the AI-IRW. Indeed, the IRW contours are spherical while the AI-IRW contours are ellipsoidal like those of the underlying Student-10 density.

\subsection{Computation time of the AI-IRW depth using both SC and MCD estimators}\label{subsec:computation}

\begin{figure}[!h]
\centering
\begin{tabular}{c}
\includegraphics[scale=0.35, trim= 1cm 0cm 0cm 0cm]{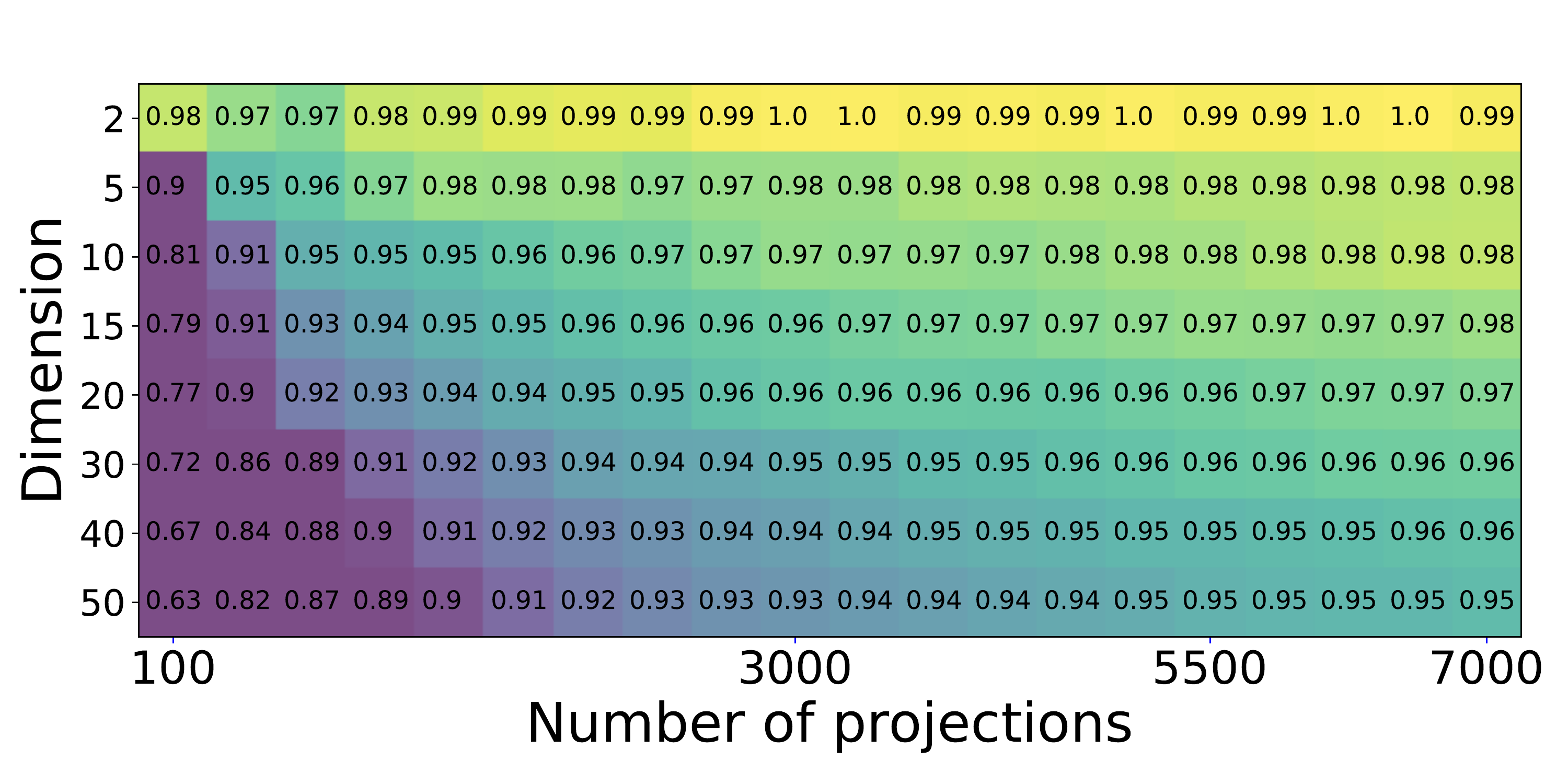}
\end{tabular}
\vspace{-0.5cm}
\caption{Kendall correlation between the approximated ranks of IRW and its true ranks depending on the number of approximating projections $m$ for a Gaussian distribution.}
\label{fig:projchoice2}
\end{figure}

\begin{figure}[!h]
\centering
\begin{tabular}{cc}
$d=2$&$d=5$\\
\includegraphics[scale=0.20, trim= 1cm 0cm 0cm 0cm]{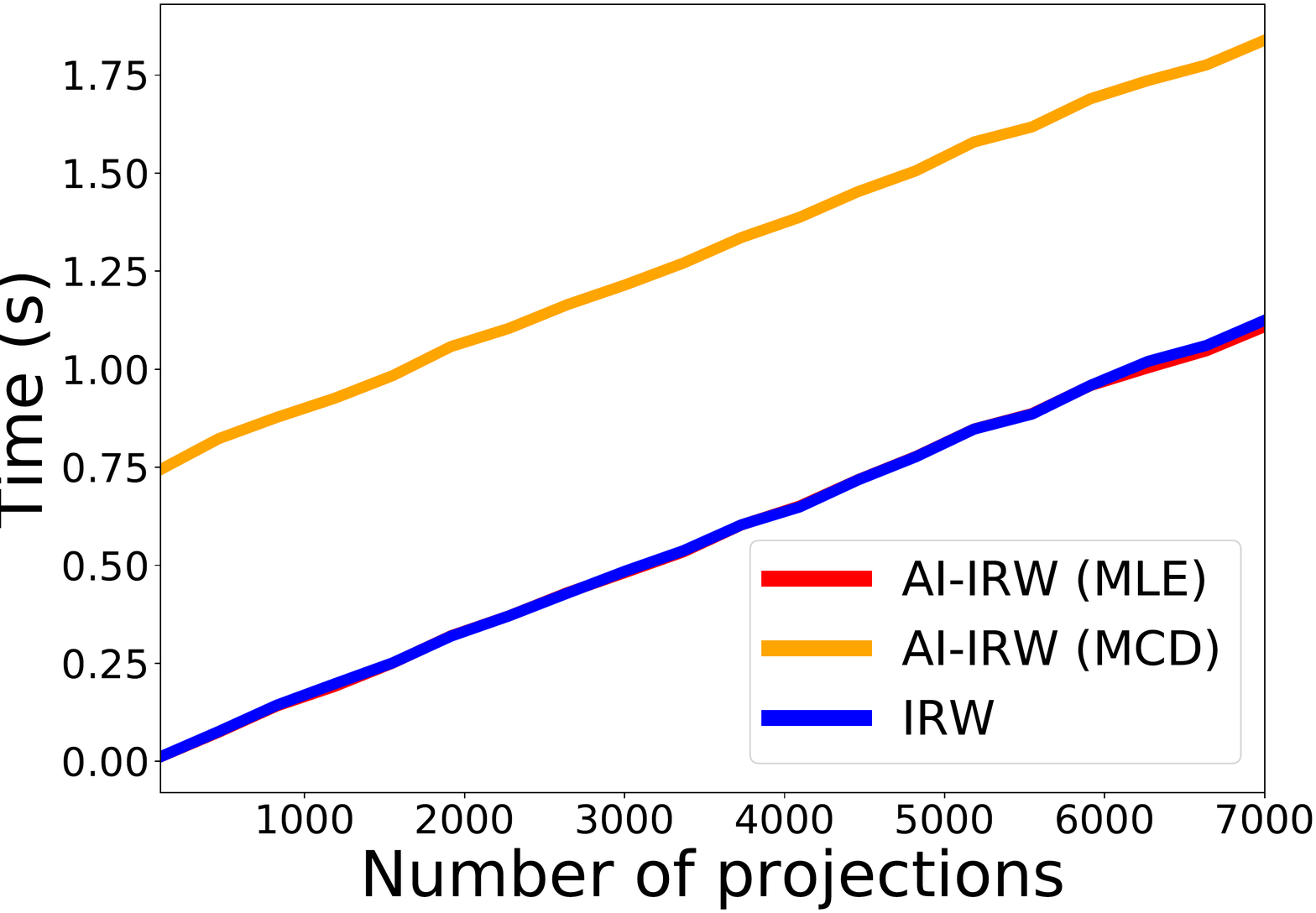}&\includegraphics[scale=0.2, trim= 1cm 0cm 0cm 0cm]{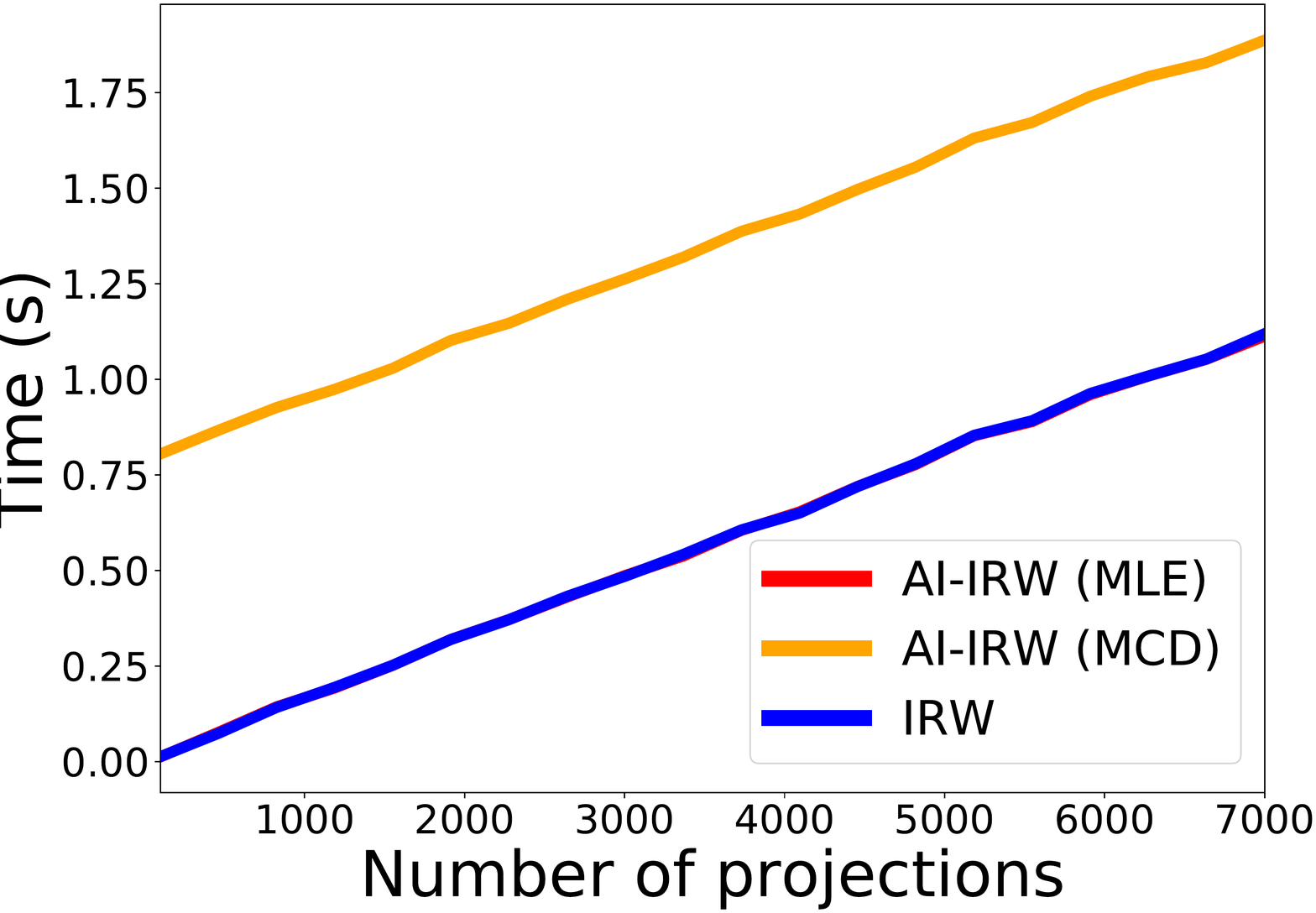}\\
$d=10$&$d=15$\\
\includegraphics[scale=0.20, trim= 1cm 0cm 0cm 0cm]{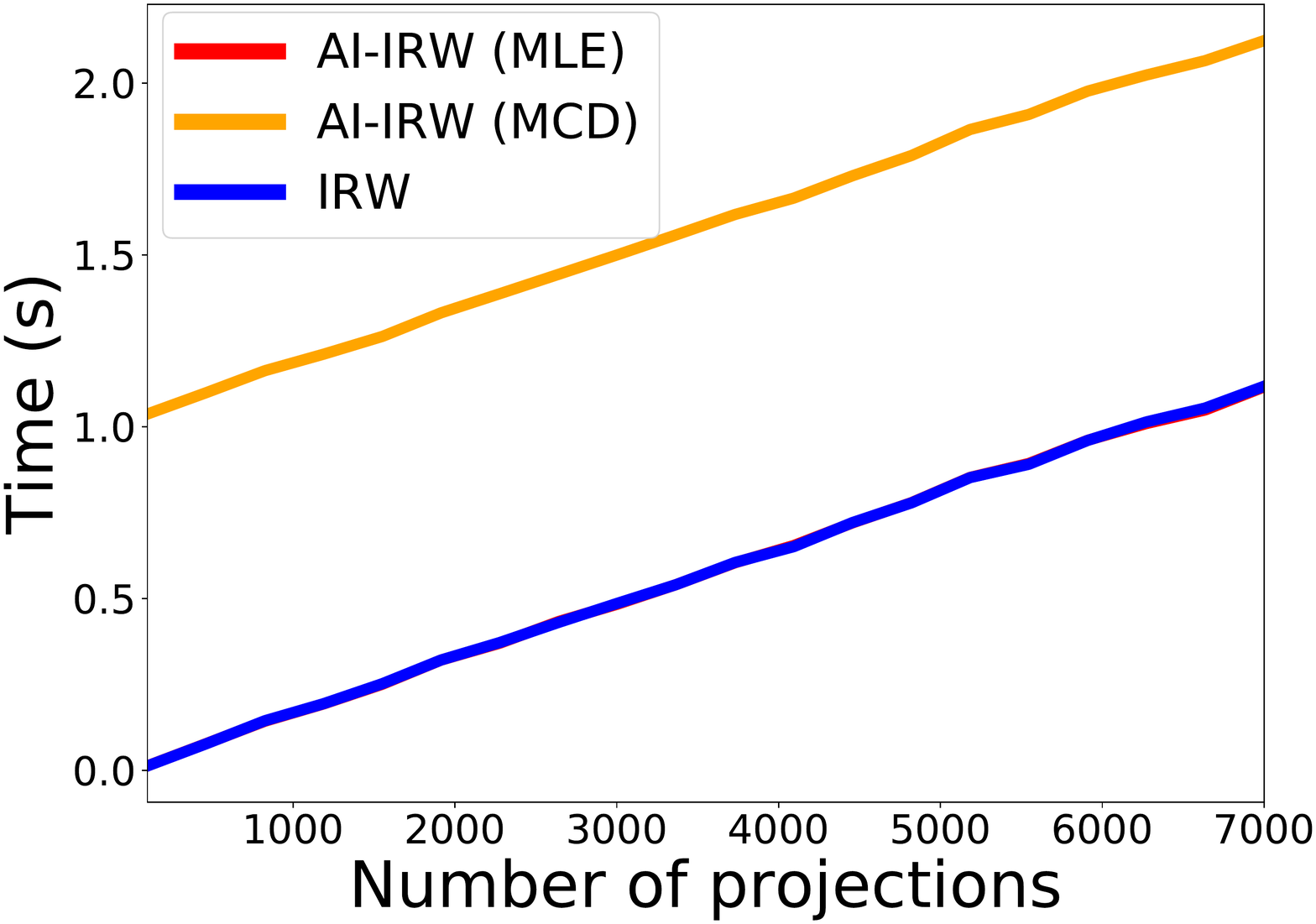}&\includegraphics[scale=0.2, trim= 1cm 0cm 0cm 0cm]{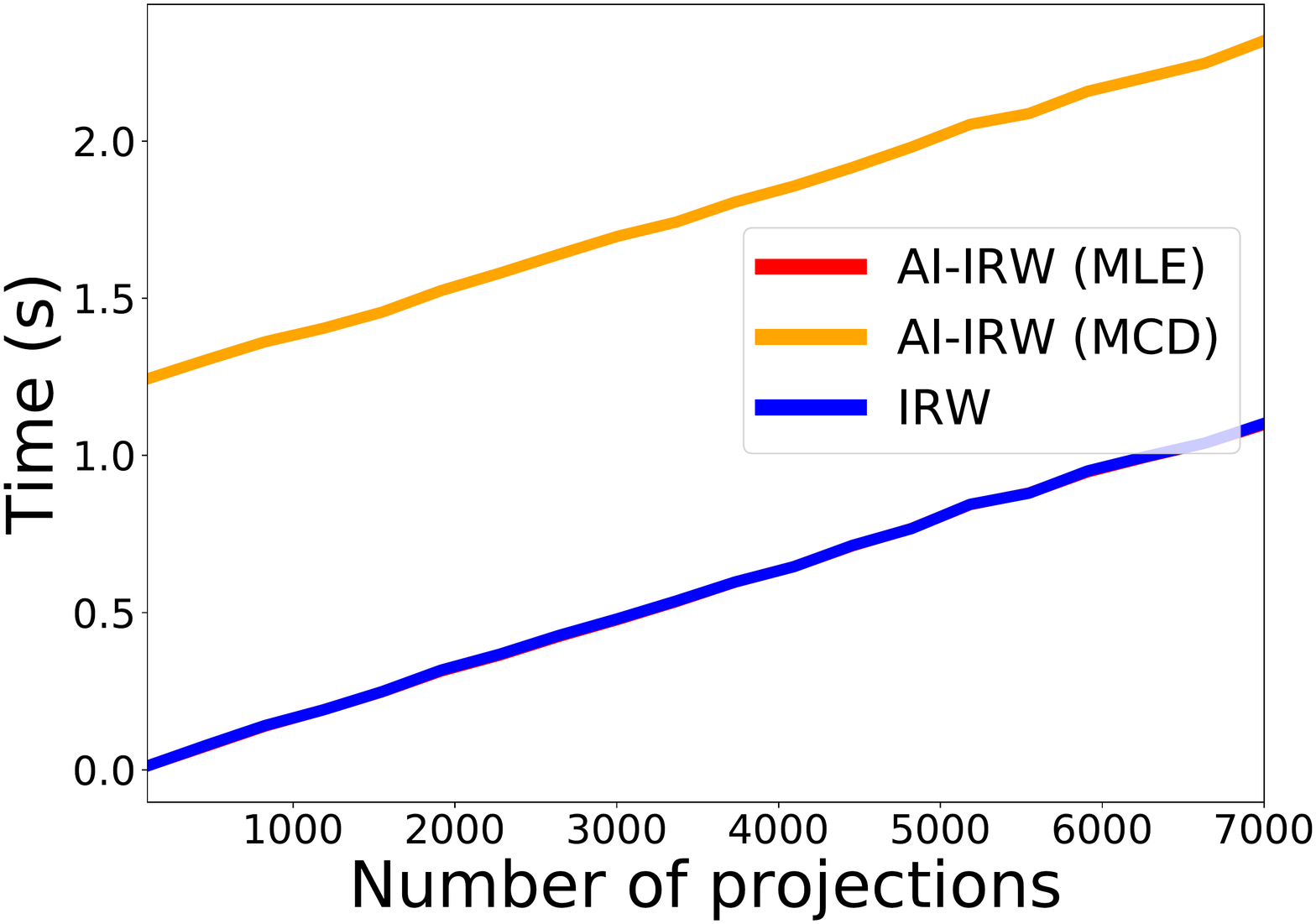}\\
$d=20$&$d=30$\\
\includegraphics[scale=0.20, trim= 1cm 0cm 0cm 0cm]{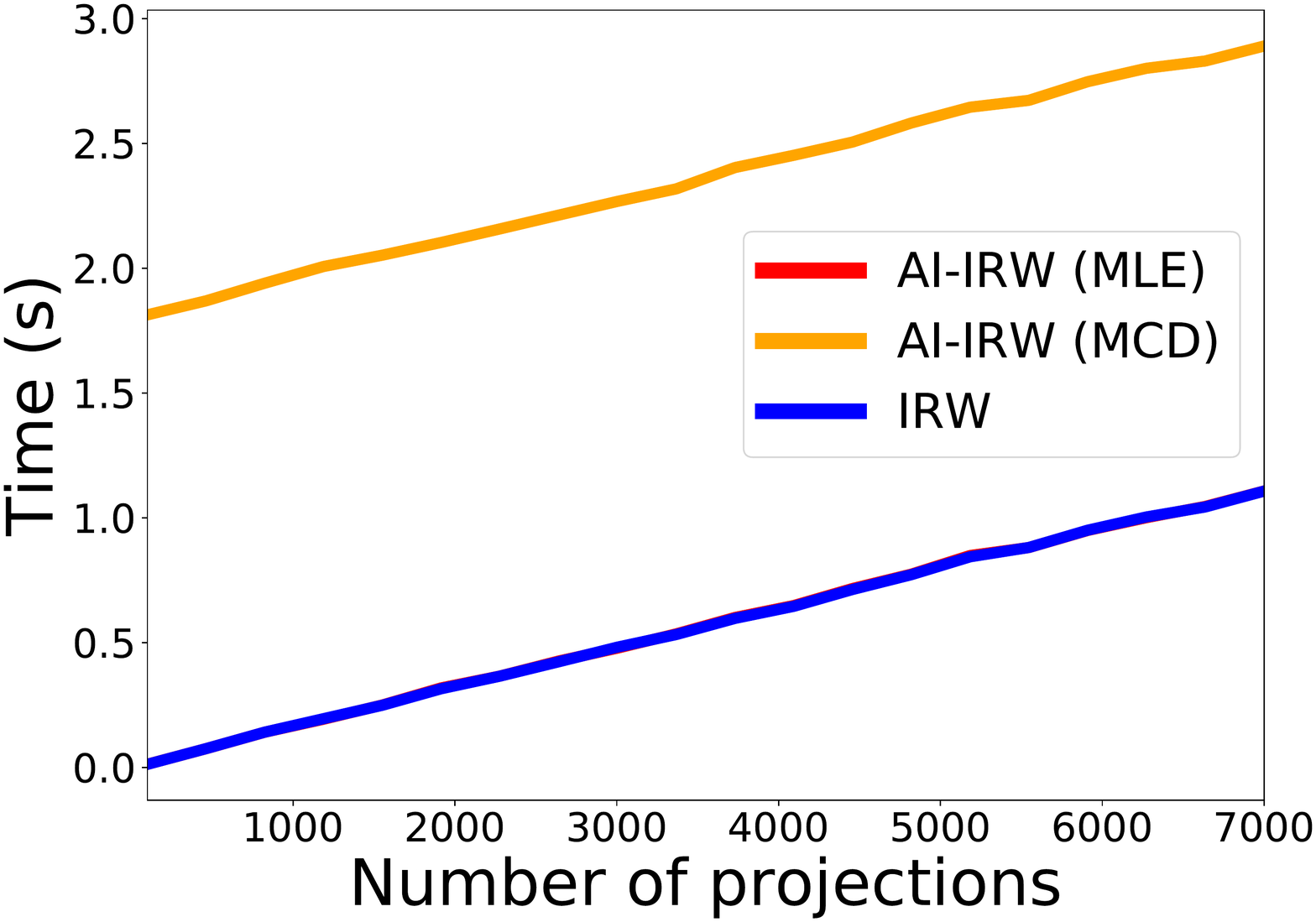}&\includegraphics[scale=0.2, trim= 1cm 0cm 0cm 0cm]{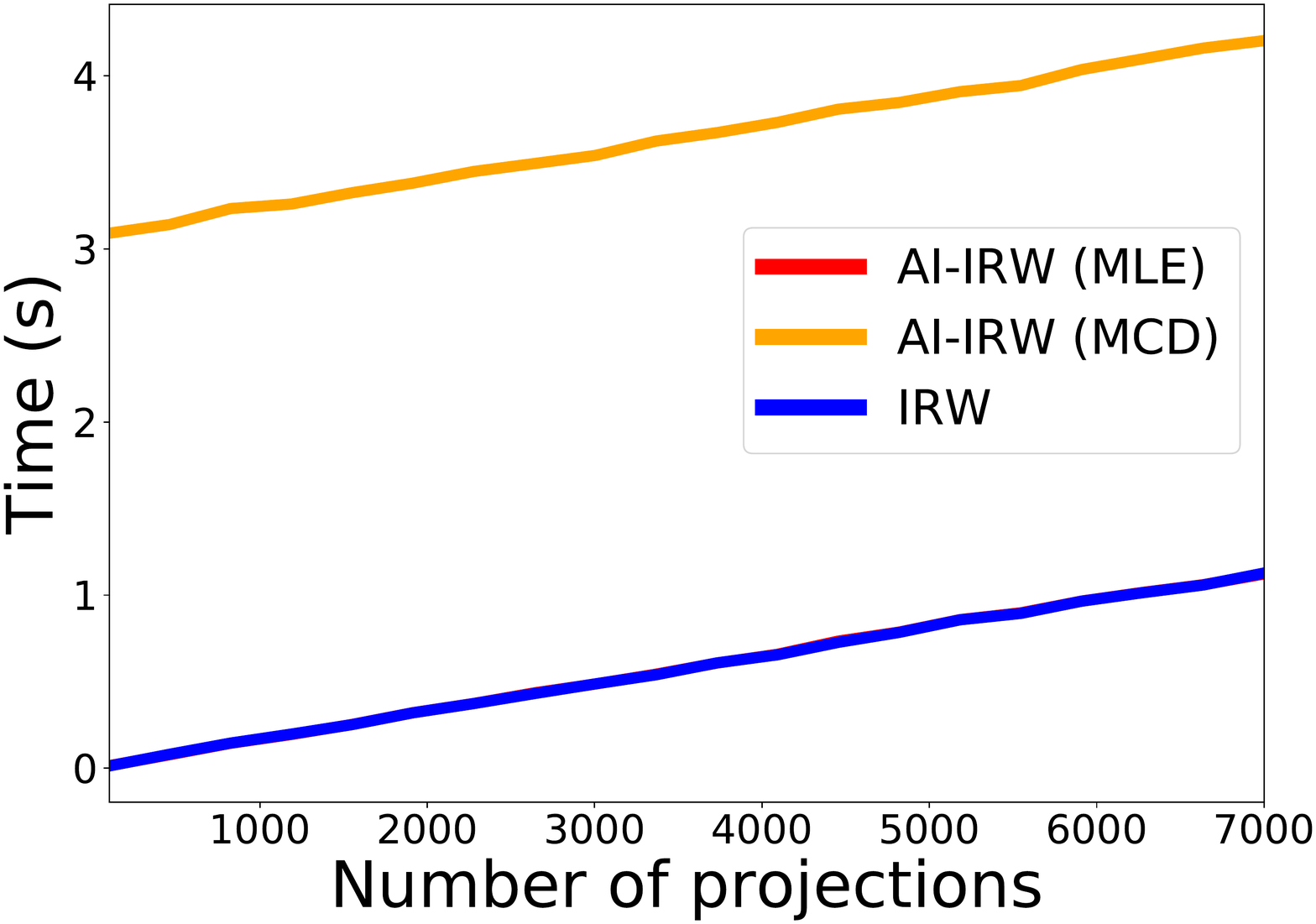}\\
$d=40$&$d=50$\\
\includegraphics[scale=0.2, trim= 1cm 0cm 0cm 0cm]{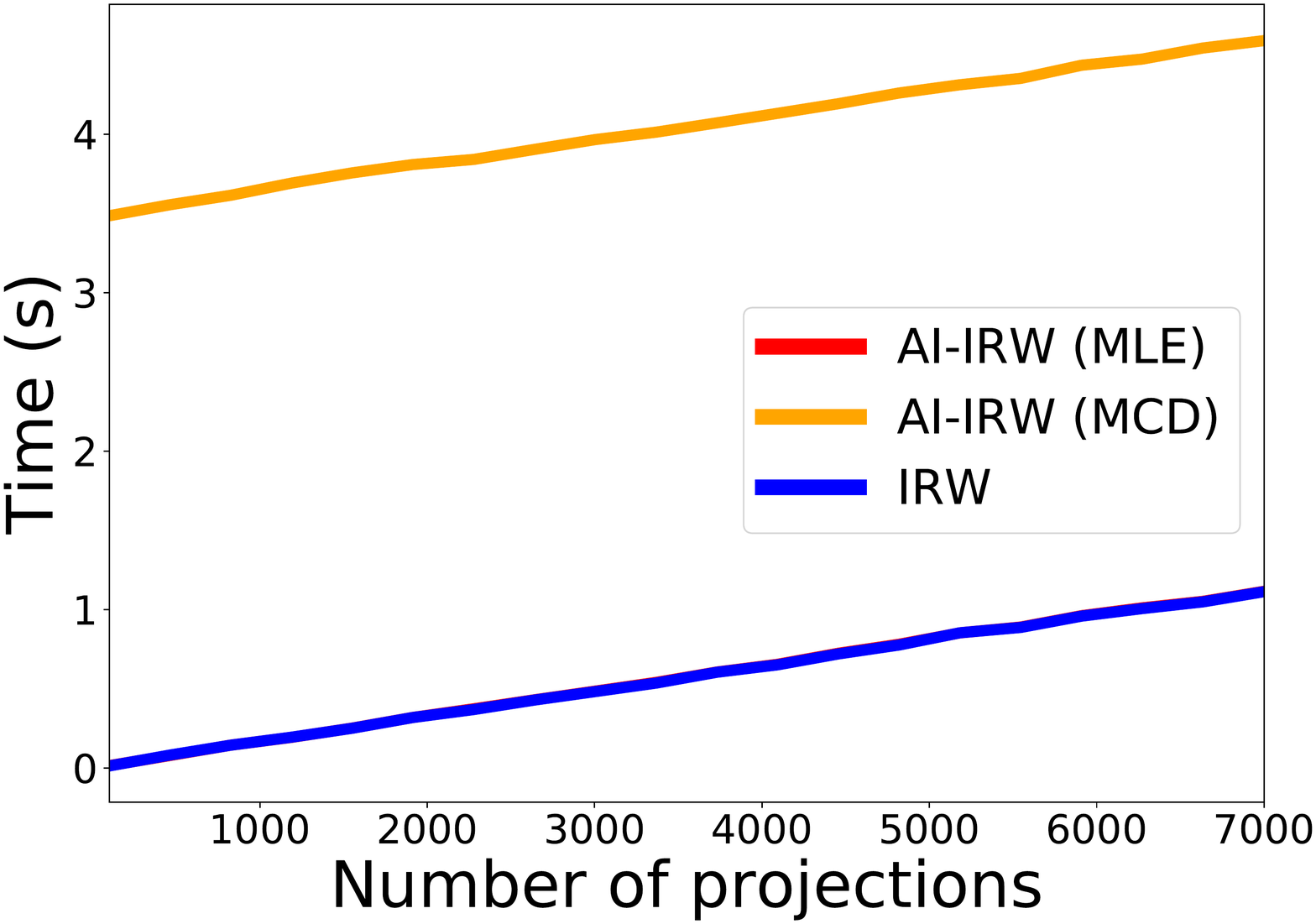}&\includegraphics[scale=0.2, trim= 1cm 0cm 0cm 0cm]{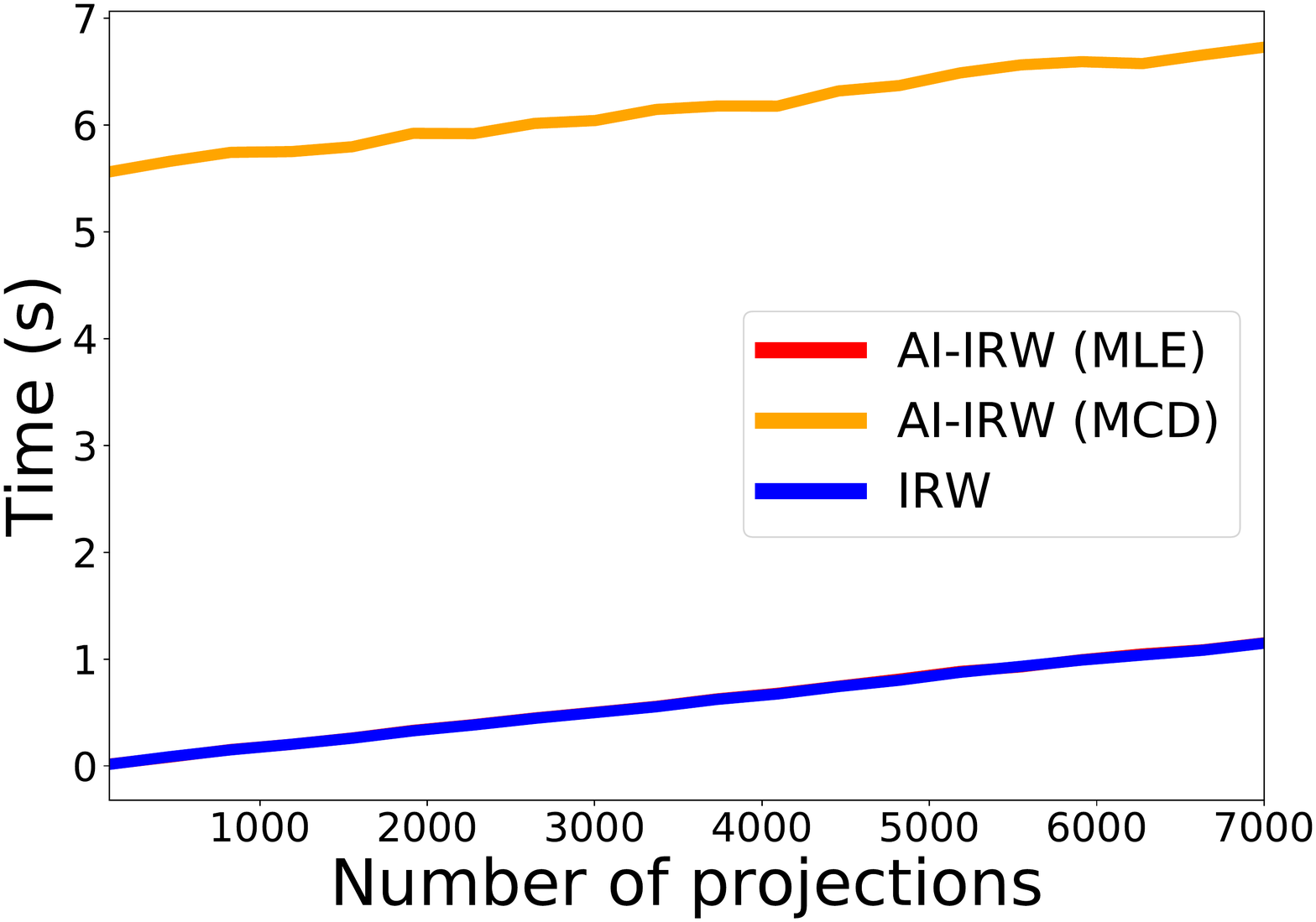}\\
\end{tabular}
\caption{Computation time of the AI-IRW depth using both SC and MCD estimators and the IRW depth depending on the number of projections for various dimensions. AI-IRW and IRW have the same computation time since the computation of the sample covariance matrix is negligible w.r.t. the computation of the IRW depth.}
\label{fig:computation}
\end{figure}

Results related to the figure~\ref{fig:projchoice} for the IRW depth are depicted in Figure~\ref{fig:projchoice2}. Kendall correlations are close to those of AI-IRW  with a slight advantage to the IRW depth as expected due to the presence of an additional covariance estimate term. In addition, computation times related to the first experiment of Section~\ref{sec:num} are displayed in Figure~\ref{fig:computation} for the AI-IRW depth using both SC and MCD estimators as well as the IRW depth. 

\clearpage
\subsection{Anomaly Detection}
\subsubsection{Anomaly Detection: a Comparison on a Toy Data Set}

\begin{figure}[!h]
\begin{center}
\begin{tabular}{cc}
\includegraphics[scale=0.55, trim= 0cm 0cm 0cm 0cm]{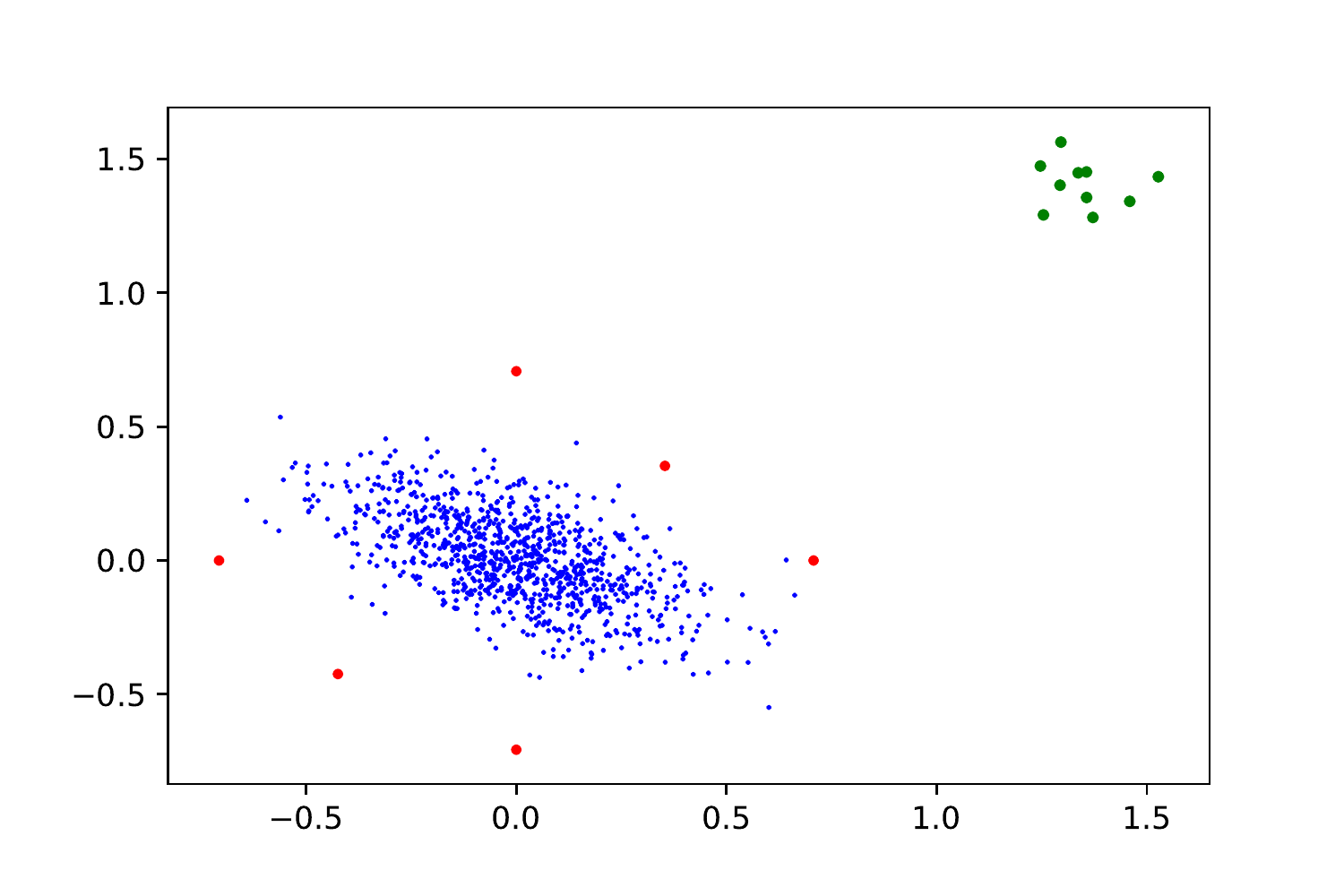}&
\includegraphics[scale=0.55, trim=0cm 0cm 0cm 0cm]{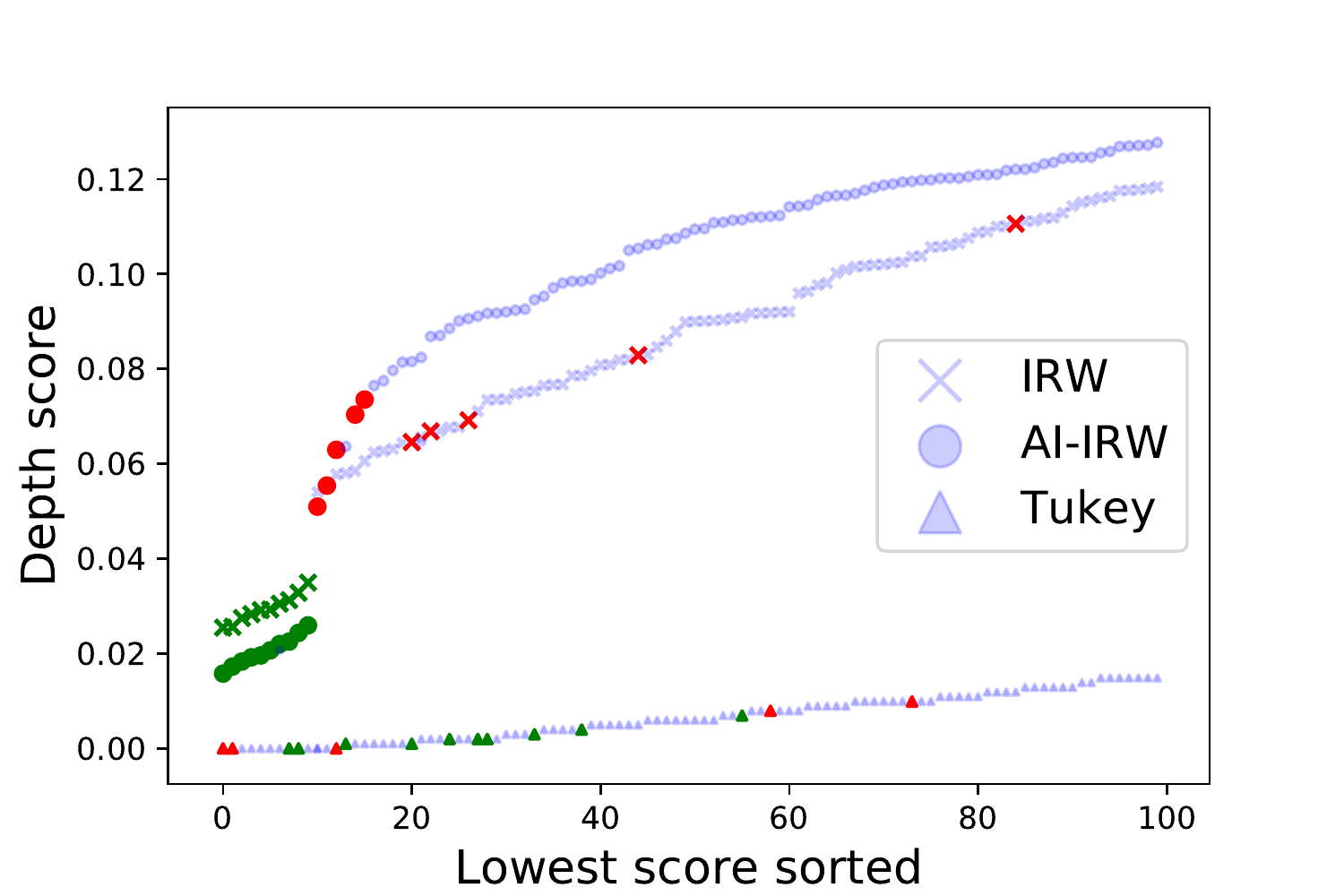}
\end{tabular}
\end{center}
\caption{Toy data set with outliers (left) and the AI-IRW, IRW and Tukey sorted scores (right).}
\label{fig:toy}
\end{figure}

In this part, a comparison between AI-IRW, IRW and the halfspace depth is provided. To conduct this experiment, we construct a toy contaminated data set (see Figure~\ref{fig:toy}, left)  where aggregated outliers (green points) and some independent outliers (red points) are added to 1000 points stemming from a 2-dimensional \emph{Gaussian} distribution. The 100 lowest scores are depicted (see Figure~\ref{fig:toy}, right) for the three benchmarked data depths. Results show that  AI-IRW is able to assign the lowest depth to these anomalies while IRW and Tukey both fail to identify them.

\subsubsection{Computation Time of Benchmarked Anomaly Detection Methods}\label{add:anom}

Here we provide additional information of the benchmarked data sets used in Section~\ref{sec:num} as well as the computation time used to perform the anomaly detection benchmark. Results are displayed in Table~\ref{computation}. AI-IRW, IRW, HM and Tukey are implemented from scratch in python using \texttt{numpy} python library. Isolation Forest implementation comes from \texttt{scikit-learn} python library \citep{sklearn} while the AutoEncoder implementation comes from \texttt{pyod} python library \citep{zhao2019pyod}. All the computations are done on a computer with 3.2 GHz Intel processor with 32 GB of RAM.

{\renewcommand{\arraystretch}{1} 
{\setlength{\tabcolsep}{0.12cm}
\begin{table}[!h]
\begin{tabular}{|c||c|c|c||c|c|}
\hline
 & $n$ & $d$ & \% of anomaly & $\hat{\gamma}~{\scriptscriptstyle (\times 0.01)}$ & $\hat{\varepsilon}~{\scriptscriptstyle (\times 0.01)}$ \\
 \hline
Ecoli & 195 & 5&26 & 0.3 & 0.2 \\ 
\hline
Shuttle & 49097 & 9& 7 & 9 & 5.7\\
\hline
Mulcross & 262144 & 4 & 10 & 100 & 10$^{\text{-10}}$\\ 
\hline
Thyroid &3772 &6 &2.5 & 0.01 & 0.1\\
\hline
Wine & 129& 13& 7.7 & 0.9 & 0.9\\ 
\hline
Http& 567479& 3& 0.4 & 19 & 2.9\\ 
\hline
Smtp& 95156& 3& 0.03 & 3.9 & 36\\
\hline
Breastw & 683& 9& 35 & 80 & 20\\
\hline
Musk& 3062 & 166& 3.2 & 9.4 & 6\\
\hline
Satimage & 5803 & 36& 1.2 & 283 & 2.6\\
\hline
\end{tabular}
\hfill
\begin{tabular}{|c||c|c|c|c|c|c|}
 \hline
&AI-IRW&IRW& HM& T& IF & AE\\
 \hline
Ecoli& 0.04& 0.005& 0.02& 0.005& 0.13& 9\\
 \hline
 Shuttle& 20& 6.8& 1.5& 6.8& 1.4& 469  \\
 \hline
 Mulcross& 75& 27& 6.2& 27& 5.9 & 2383\\
 \hline
 Thyroid &  1 & 0.21& 0.2& 0.2& 0.18 & 42\\
 \hline
 Wine& 0.05& 0.01& 0.06& 0.008& 0.12 & 8.1\\
 \hline
 Http & 97 & 45& 11& 45& 11 &  5197 \\
 \hline
 Smtp&  22 & 4.5 & 1 & 4.5 & 1.88 & 903\\
 \hline
 Breastw& 0.46& 0.04& 0.06&  0.04& 0.14 & 17.2 \\
 \hline
 Musk& 20.5&  5.23&  2.5& 5.2&0.43 & 103 \\
 \hline
 Satimage& 6.3& 2.63& 0.9& 2.6& 0.31 & 76  \\
 \hline 
\end{tabular}
\caption{Left: Data sets considered for the performance comparison: $n$ is the number of instances, $d$ is the number of attributes, $\hat{\gamma}$ and $\hat{\varepsilon}$ are the eigengap and the smallest eigenvalue of the SC estimator respectively (left). Right: Computation time of benchmarked anomaly detection methods in seconds.}
\label{computation}
\end{table}}}

\subsection{Exploring AI-IRW with the MCD Estimator}

In this section, we investigate the quality of the approximation as well as  the robustness of the AI-IRW depth using both SC and MCD estimators with two experiments.

\subsubsection{Approximation and robustness} 
\begin{figure}[h]
\centering
\begin{tabular}{cc}
\includegraphics[scale=0.27, trim= 0cm 0cm 0cm 0cm]{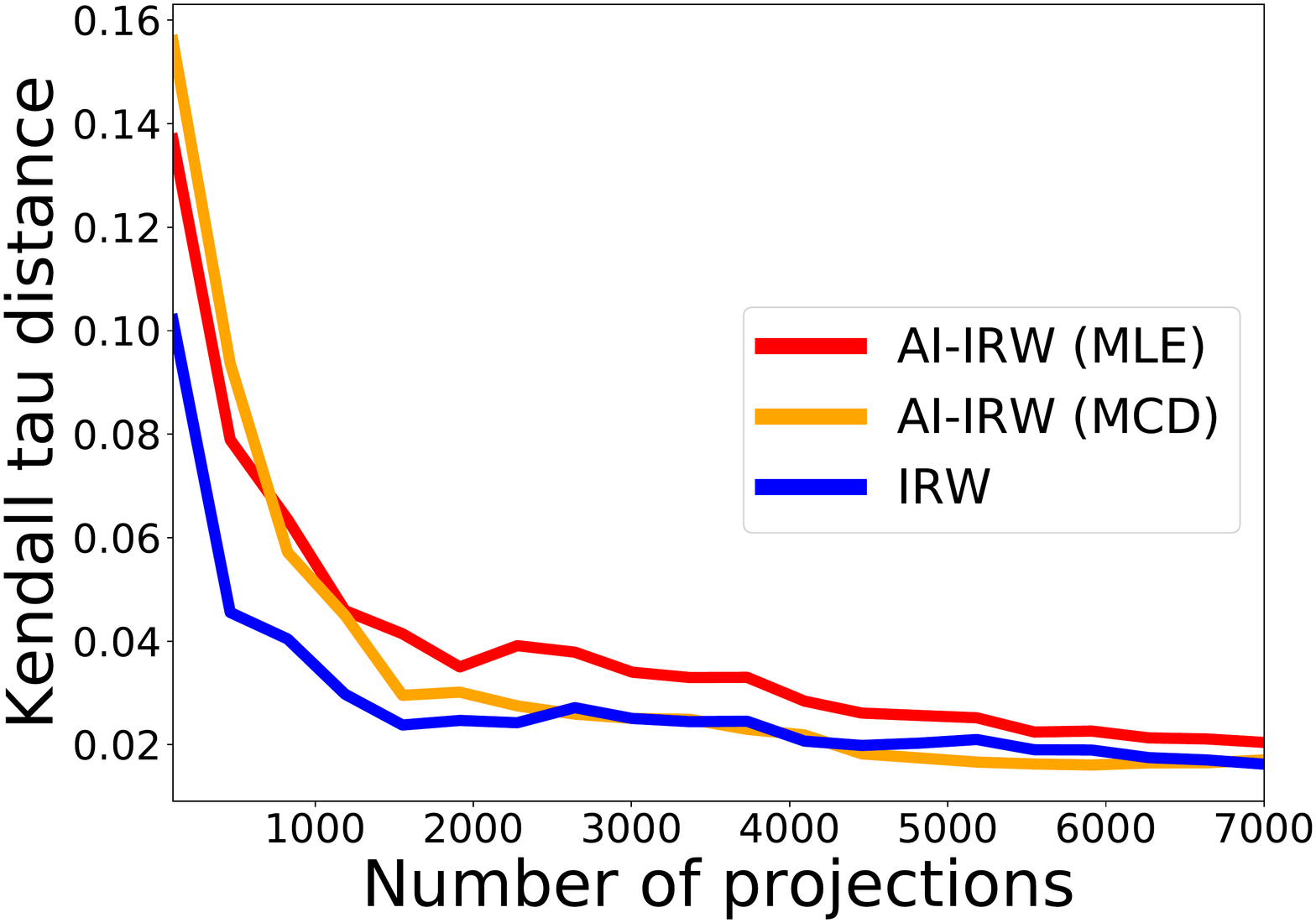} &  \includegraphics[scale=0.27, trim= 0cm 0cm 0cm 0cm]{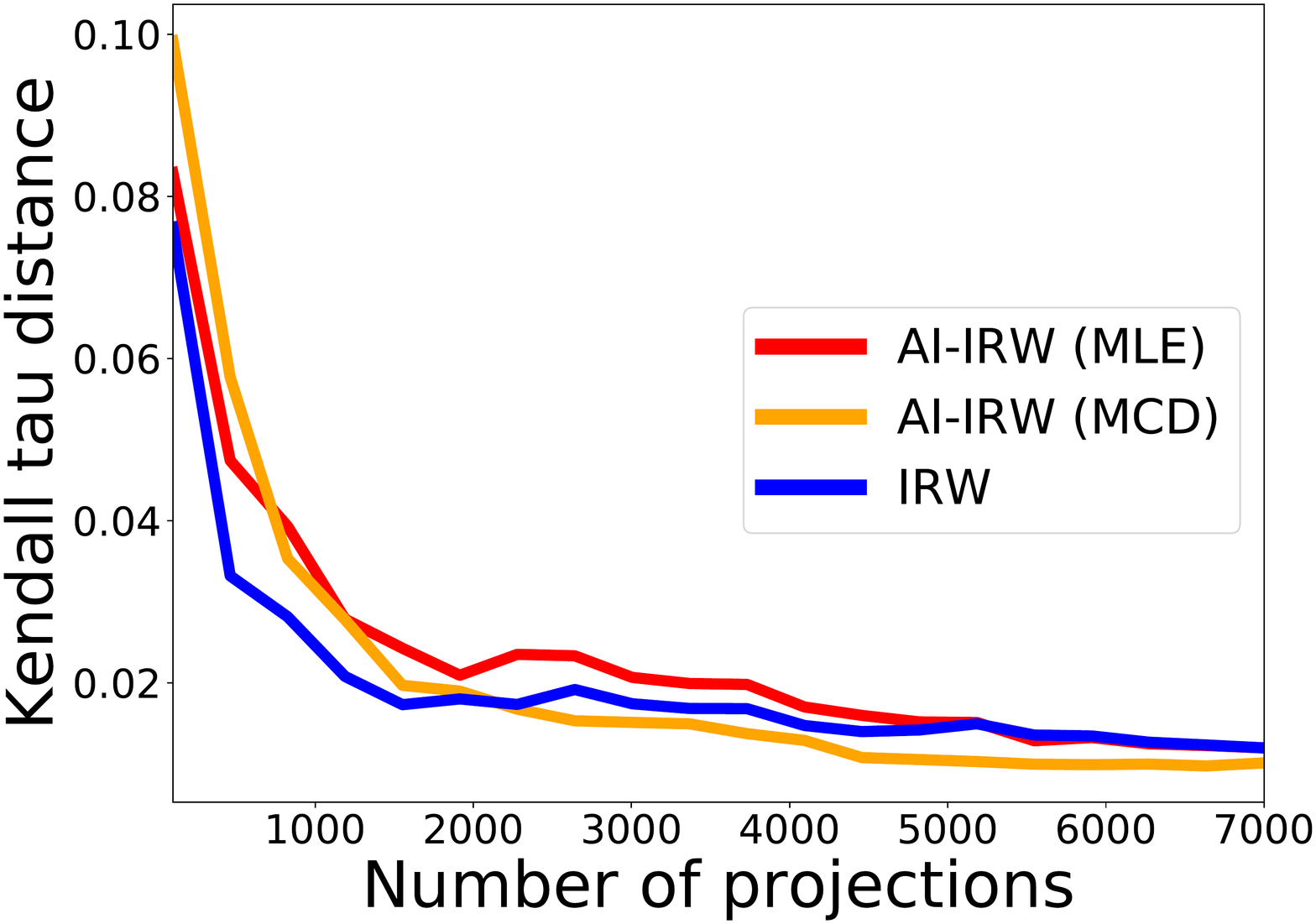}
\end{tabular}
\caption{Coherence of the returned rank measured by Kendall $\tau$ distance depending on the number of approximating projections for Gaussian (left) and Student-3 (right) distributions for AI-IRW (using SC and MCD estimates) and IRW.}
\label{fig:rank}
\end{figure}

The first experiment is conducted as follows. The accuracy of Monte Carlo approximation, depending on the number $m$ of random directions uniformly sampled, is evaluated for the empirical versions of the AI-IRW depth using SC and MCD estimators as well as the IRW depth. The experiment is based on samples of size $n=1000$ drawn from the multivariate standard Gaussian distribution (standard, so that non affine invariant depth are not disadvantaged) in dimension $d=5$. The classical Kendall $\tau$ distance, given by
$$d_{\tau}(\sigma, \sigma ')=\frac{2}{n(n-1)} \sum_{i<j} \mathbb{I}_{\{(\sigma (i)-\sigma (j))(\sigma '(i)-\sigma '(j))<0 \}},$$
for all permutations $\sigma$ and $\sigma'$ of the index set $\{1,\; \ldots,\; n\}$, 
 is used to measure the deviation between the ranks induced by the ‘‘true'' depth (approximated with $m=200000$ projections since there exists no closed-form) and those defined by the Monte Carlo approximation of the sampling version.
The averaged Kendall $\tau$'s  (over $10$ runs), that correspond to one minus the Kendall correlations,  are displayed in Figure~\ref{fig:rank}. One observes that the approximate empirical AI-IRW depth is not affected by the covariance estimation step, its behavior is similar to that of the approximate empirical IRW depth for the Gaussian distribution when using both covariance estimators. On the other hand, a slight advantage is awarded to MCD under the heavy-tailed Student-3 model.

\subsubsection{Robustness w.r.t. increasing proportion of outliers} 
\begin{figure}[h]
\centering
\begin{tabular}{cc}
\includegraphics[scale=0.27, trim= 0cm 0cm 0cm 0cm]{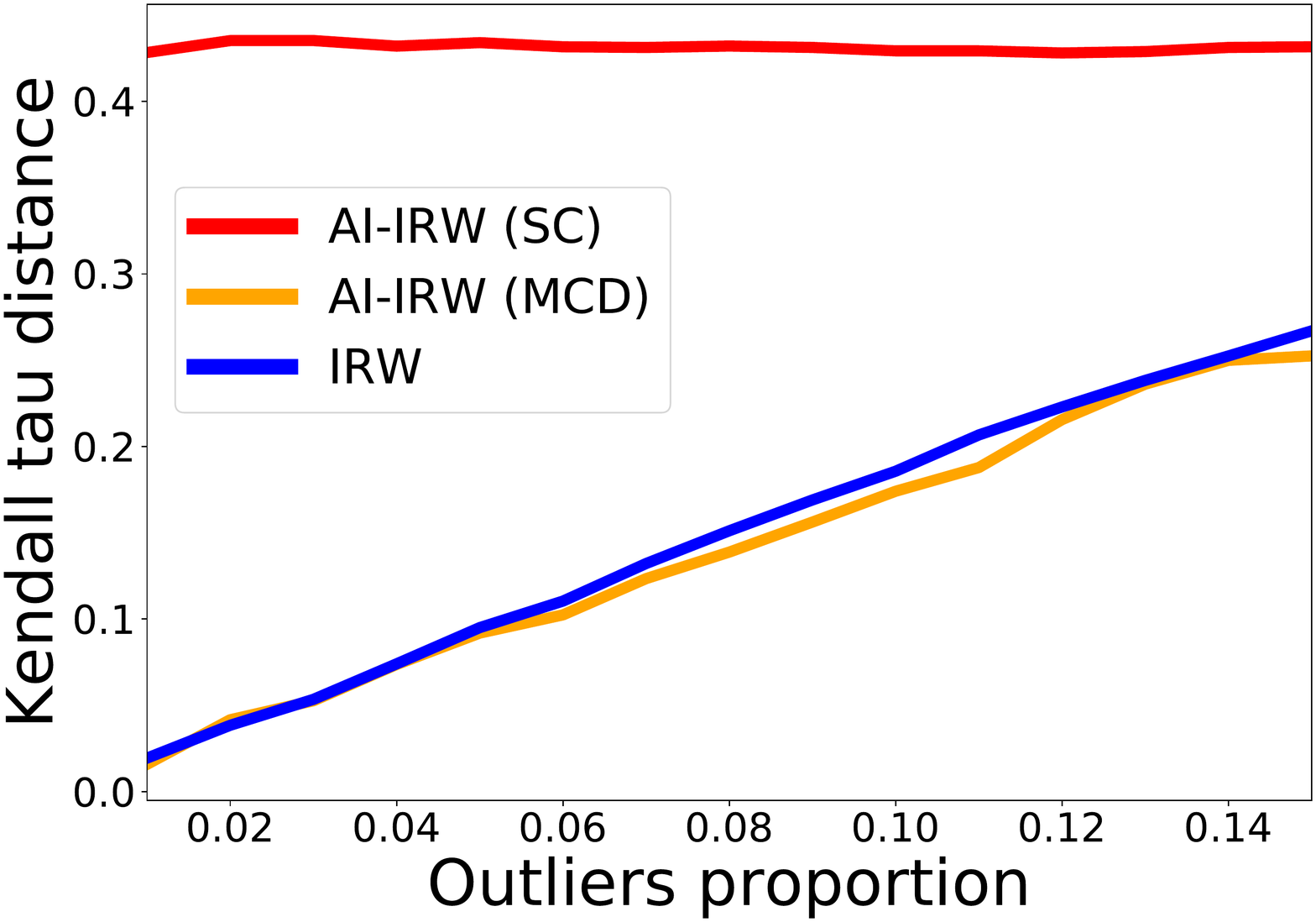} &  \includegraphics[scale=0.27, trim= 0cm 0cm 0cm 0cm]{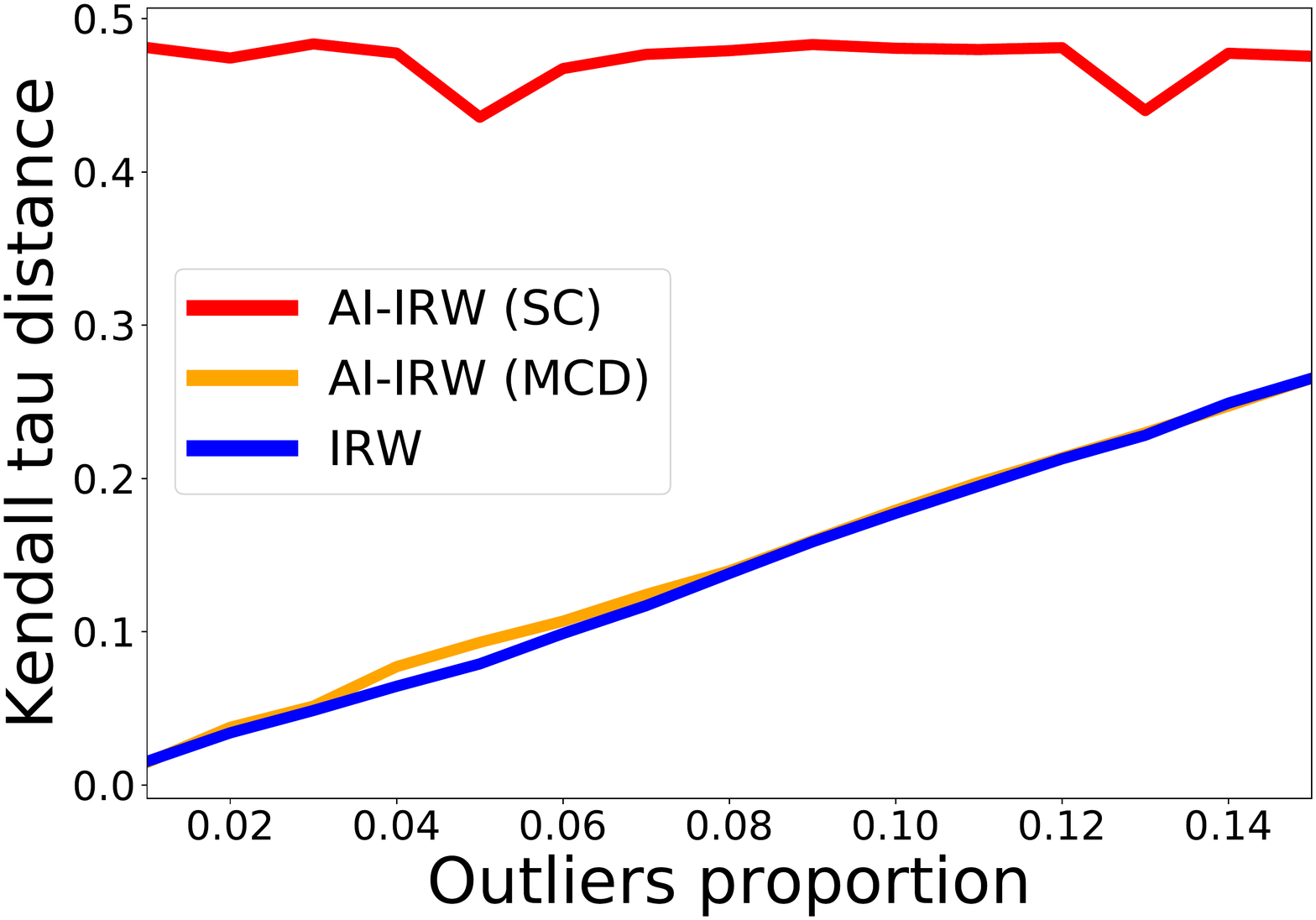}
\end{tabular}
\caption{Coherence of the returned rank measured by Kendall $\tau$ depending on outliers proportion for Student-3 (left) and Gaussian (right) distributions for AI-IRW (using SC and MCD estimates) and IRW.}
\label{fig:robustness}
\end{figure}

In the second experiment, we examine the robustness of the returned ordering. It is based on the construction of two contaminated data sets from samples of size $n=100$ drawn from the multivariate standard Gaussian distribution (standard, so that non affine invariant depths are not disadvantaged) in dimension $d=2$. To build corrupted data set, the two following contaminated models are used. The first is based on adding ‘‘isolated outliers'' where each of them is defined as $(0,a)$ where $a$ is sampled uniformly between $[4,400]$. The second is based on adding ‘‘aggregated outliers'' by randomly and uniformly drawing a location $b$ in $[4,400]$ and then drawing anomalies following the Gaussian distribution $\mathcal{N}(\mathbf{b}, \mathcal{I}_2)$ where $\mathbf{b}$ is the vector $(b,b)$. Therefore, each data set is constructed as follows: a proportion of outliers  $\alpha\in [0,0.15]$ is added to the normal data, represented by the standard Gaussian distribution, following one of the two aforementioned contamination models and thus yields two settings. The AI-IRW depth using SC and MCD estimators as well as the IRW depth are computed on these contaminated data sets. The Kendall $\tau$ distance is used to measure the deviation between the ‘‘true'' ranks that are computed on samples without corruption and those computed on samples with corruption w.r.t. a proportion of anomalies $\alpha$. The averaged Kendall $\tau$'s  (over $100$ runs) are displayed in Figure~\ref{fig:robustness}. As expected, results show that the MCD estimator provides robustness to the AI-IRW depth while the sample covariance estimator breaks down after only $1\%$ of anomalies. Interestingly, the MCD estimator does not bring more robustness than the underlying robustness of the IRW depth. It highlights somehow a ‘‘worst case'' robustness between the estimator of the covariance matrix and the underlying IRW depth which is reached by the latter.

\subsection{Variance of AI-IRW Score}
\subsubsection{Variance w.r.t.  Sample Realizations}\label{var}

\begin{figure}[h]
\begin{center}
\begin{tabular}{cc}
\includegraphics[scale=0.25, trim= 0cm 0cm 0cm 0cm]{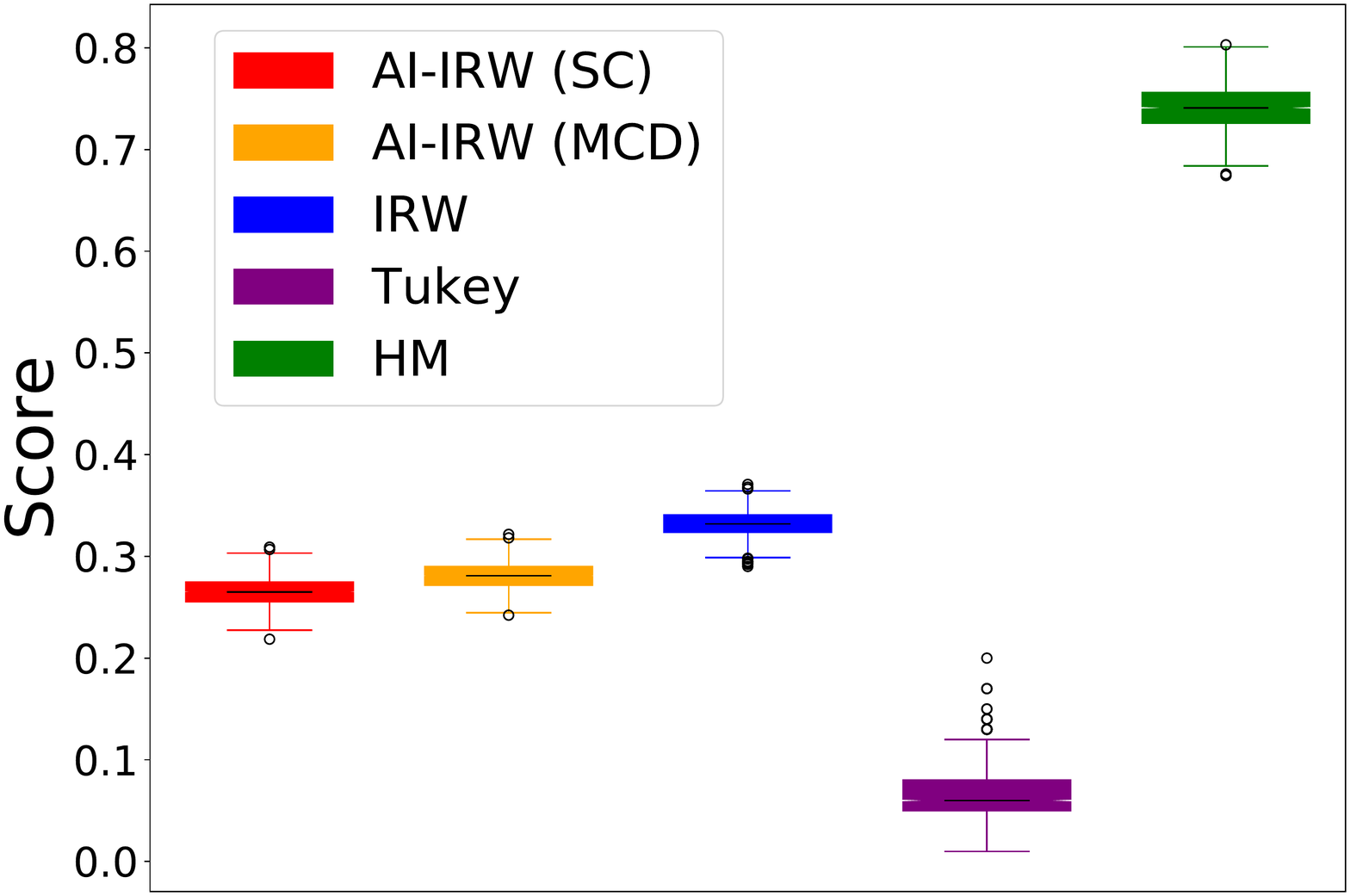}&\includegraphics[scale=0.25, trim= 0cm 0cm 0cm 0cm]{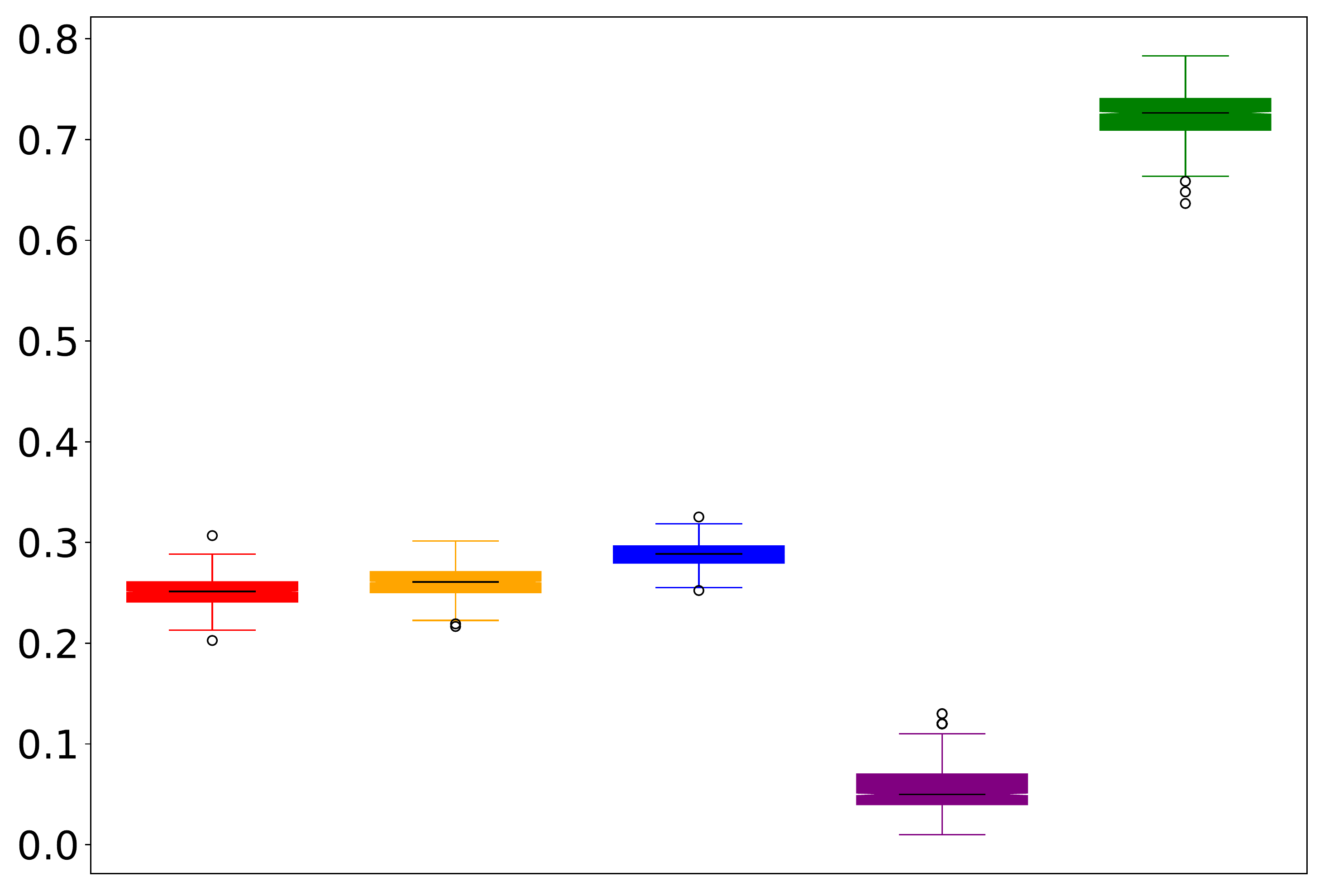}
\end{tabular}
\end{center}
\caption{Variance of the score of $x_1,x_2$ (from left to right) over 1000 repetitions for the AI-IRW, IRW, halfspace mass (HM) and halfspace (Tukey) depths. }
\label{fig:variance}
\end{figure}

We compare the stability of the approximation estimator AI-IRW measuring its variance. For 100 points stemming from a 10-dimensional Gaussian distribution with zero mean and covariance matrix drawn from the Wishart distribution (with parameters $(d,\mathcal{I}_d)$) on the space of definite matrices, the variance of the returned score is computed on two points, denoted by $x_1$ and $x_2$, drawn randomly from the 100 points previous points. The score is computed for AI-IRW, IRW, halfspace mass and halfspace depths each approximated using $m=100$ directions. Figure~\ref{fig:variance} illustrates that (1) no additional variance is introduced by the affine-invariant version, (2) closeness of the three scores (due to absence of correlation), as well as (3) their higher concentrations compared to halfspace mass and halfspace depth.

\subsubsection{Variance  w.r.t. Noisy Directions}

\begin{figure}[!h]
\begin{center}
\begin{tabular}{ccc}
\includegraphics[scale=0.25, trim= 0cm 0cm 0cm 0cm]{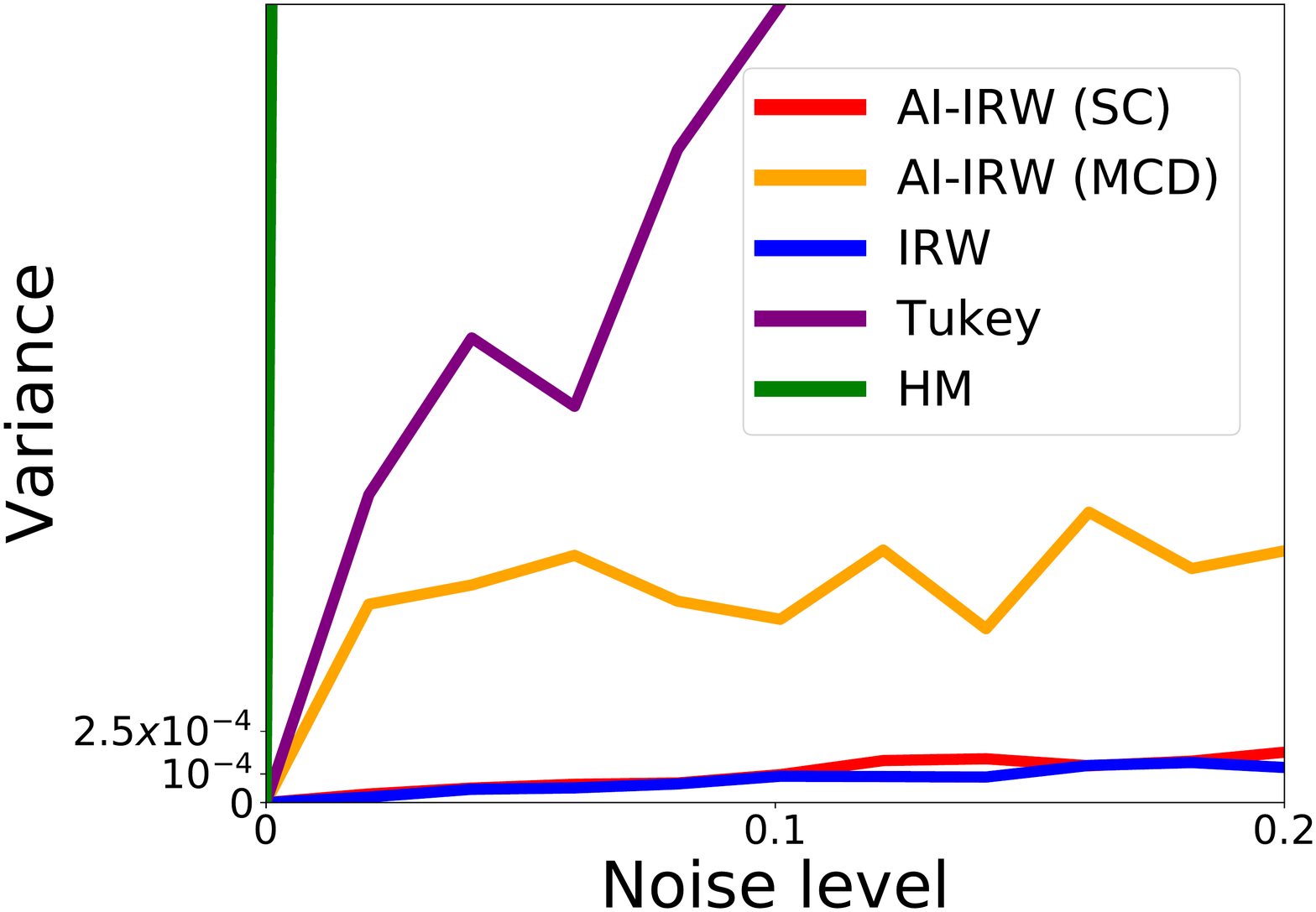}&\includegraphics[scale=0.25, trim=0cm 0cm 0cm 0cm]{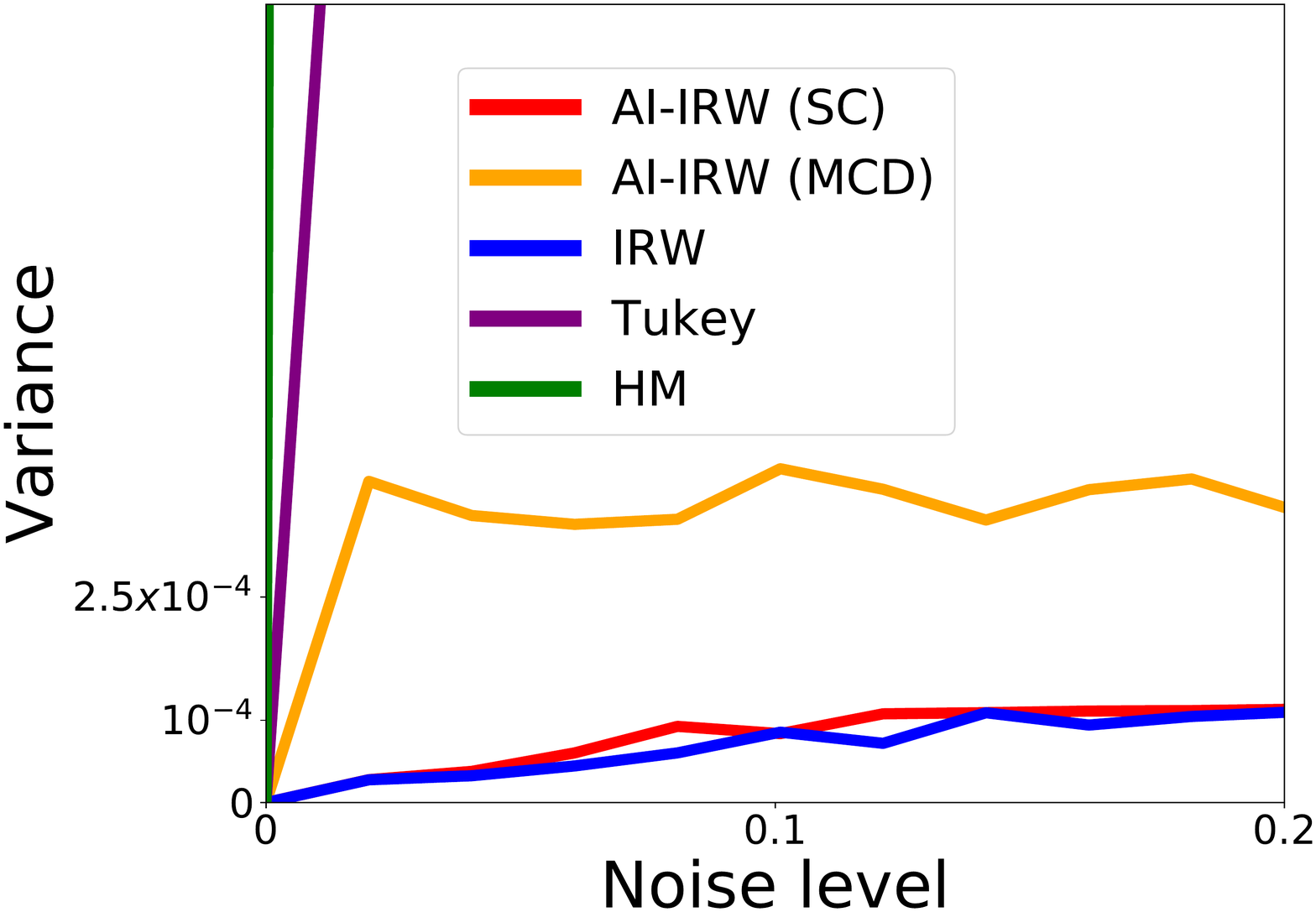}
\end{tabular}
\end{center}
\caption{Variance of the score of $x_1,x_2$ (from left to right) over the noise level induced in sampled directions with 1000 repetitions for the AI-IRW, IRW,  Tukey depth.}
\label{fig:noise}
\end{figure}

The experiment in Section~\ref{var} is repeated with different level of Gaussian noise that are added to sampled directions, i.e.  $U= \frac{W+\varepsilon \mathcal{N}(\mathbf{0}, \mathcal{I}_d) }{||W+\varepsilon \mathcal{N}(\mathbf{0}, \mathcal{I}_d ||}$. This experiment is conducted with AI-IRW, IRW, HM and halfspace depth using $m=100$ sampled directions. The root mean square variance (over 100 repetitions) between the returned score and the  original score (without noise) are computed for $x_1,x_2$ (same as those in  Section~\ref{var}), see Figure ~\ref{fig:noise}. Results show that AI-IRW (using the SC estimator) shares  very few differences with IRW  while the superiority of AI-IRW (and IRW) over the existing methods depth such as haflspace and halfspace mass is highlighted. 

\end{document}